\documentclass[10pt,twocolumn,letterpaper]{article}

\usepackage{cvpr}
\usepackage{times}
\usepackage{epsfig}
\usepackage{caption}
 \usepackage{url}
\usepackage{xcolor}
\usepackage{soul}
\usepackage[percent]{overpic}
\usepackage[numbers,sort]{natbib}
\usepackage{graphicx}
\usepackage{amsmath}
\usepackage[all=normal,paragraphs=tight]{savetrees}
\usepackage{float}
\usepackage{amssymb}

\usepackage{booktabs,tabularx} 

\usepackage{tikz}
\usepackage{verbatim}
\usetikzlibrary{calc}
\usetikzlibrary{positioning}
\usetikzlibrary{3d}

\usepackage{tikz-3dplot}
\usetikzlibrary{calc,arrows.meta,positioning,backgrounds}

\usepackage[breaklinks=true,bookmarks=false]{hyperref}

\usepackage{pdfcomment}
\newcommand{\modelname}{SynSin}

\newcolumntype{Y}{>{\raggedright\arraybackslash}X} 

\newcommand{\figref}[1]{Fig.~\ref{#1}}
\newcommand{\secref}[1]{Section~\ref{#1}}
\usepackage{subfigure}

\newcommand{\tabref}[1]{Table~\ref{#1}}

\def\mat#1{\mathchoice{\mbox{\boldmath$\displaystyle\tt#1$}}
{\mbox{\boldmath$\textstyle\tt#1$}}
{\mbox{\boldmath$\scriptstyle\tt#1$}}
{\mbox{\boldmath$\scriptscriptstyle\tt#1$}}}
\def\m#1{\protect\mat #1}

 \cvprfinalcopy 

\def\cvprPaperID{00} 

\ifcvprfinal\fi

\begin{document}

\title{\modelname{}: End-to-end View Synthesis from a Single Image}

\author{
  Olivia Wiles\textsuperscript{1}\footnotemark \hspace{5mm}
  Georgia Gkioxari\textsuperscript{2} \hspace{5mm}
  Richard Szeliski\textsuperscript{3} \hspace{5mm}
  Justin Johnson\textsuperscript{2,4} \hspace{5mm}
  \\*[3mm]
  \textsuperscript{1}University of Oxford \hspace{2mm}
  \textsuperscript{2}Facebook AI Research \hspace{2mm}
  \textsuperscript{3}Facebook \hspace{2mm}
  \textsuperscript{4}University of Michigan
}

\twocolumn[{%
	\maketitle
	\renewcommand\twocolumn[1][]{#1}%
	\begin{center}
    \vspace{-2mm}\includegraphics[width=\textwidth]{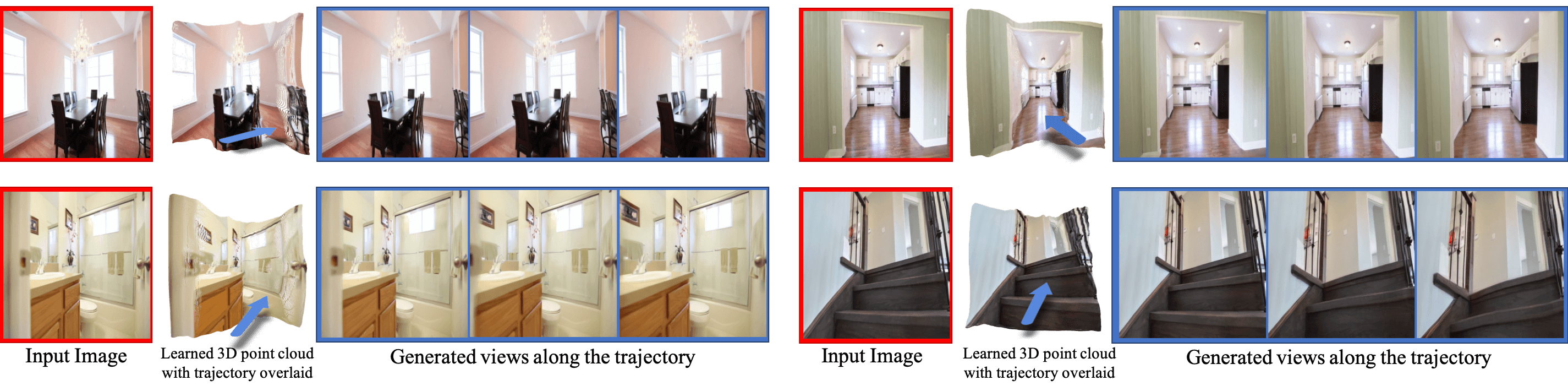}\vspace{-2mm}
		\captionof{figure}{
		{\bf End-to-end view synthesis.}
    Given a \emph{single} RGB image (\textcolor{red}{red}), \modelname{} generates images of the scene at new viewpoints (\textcolor{blue}{blue}).
    \modelname{} predicts a 3D point cloud, which is projected onto new views using our differentiable renderer; the rendered point cloud is passed to a GAN to synthesise the output image.
    \modelname{} is trained end-to-end, without 3D supervision. 
  }
		\label{fig:teaser}
	\end{center}%
}]

\begin{abstract}
\vspace{-0.6em}
View synthesis allows for the generation of new views of a scene given one or more images.
This is challenging; it requires comprehensively understanding the 3D scene from images.
As a result, current methods typically use multiple images, train on ground-truth depth, or are limited to synthetic data. 
We propose a novel end-to-end model for this task using a {\em single image} at test time; it is trained on real images without any ground-truth 3D information. 
To this end, we introduce a novel differentiable point cloud renderer that is used to transform a latent 3D point cloud of features into the target view. 
The projected features are decoded by our refinement network to inpaint missing regions and generate a realistic output image. 
The 3D component inside of our generative model allows for interpretable manipulation of the latent feature space at test time, \eg we can animate trajectories from a single image. Additionally, we can generate high resolution images and generalise to other input resolutions. We outperform baselines and prior work on the Matterport, Replica, and RealEstate10K datasets. 
\end{abstract}

\ifcvprfinal{
  \renewcommand*{\thefootnote}{\fnsymbol{footnote}}
  \footnotetext{Project page: \url{www.robots.ox.ac.uk/\~ow/synsin.html}.}
  \setcounter{footnote}{1}
  \footnotetext{Work done during an internship at Facebook AI Research.}
  \renewcommand*{\thefootnote}{\arabic{footnote}}
  \setcounter{footnote}{0}
} \fi

\section{Introduction}
\label{sec:intro}
Given an image of a scene, as in \figref{fig:teaser} (top-left), what would one see when turning left or walking forward?
We can reason that the window and the wall will extend to the left and more chairs will appear to the right.
The task of novel \emph{view synthesis} addresses these questions: 
given a view of a scene, the aim is to generate images of the scene from new viewpoints.
This task has wide applications in image editing, animating still photographs or viewing RGB images in 3D.
To unlock these applications 
for any input image,
our goal is to perform view synthesis in \emph{complex, real-world scenes} using only \emph{a single input image}.

View synthesis is challenging, as it requires comprehensive scene understanding.
Specifically, successful view synthesis requires understanding both the \textbf{3D structure} and the \textbf{semantics} of the input image.
Modelling 3D structure is important for capturing the relative motion of visible objects under a view transform.
For example in \figref{fig:teaser} (bottom-left), the sink is closer than the shower and thus shifts more as we change viewpoints.
Understanding semantics is necessary for synthesising plausible completions of partially visible objects, \eg the chair in \figref{fig:teaser} (top-left).

One way to overcome these challenges is to relax the single-image constraint and use \emph{multiple} views to reconstruct 3D scene geometry~\cite{Debevec96,Seitz06,Zitnick04,Fitzgibbon05}.
This also simplifies semantic modelling, as fewer positions will be occluded from all views.
Recent methods~\cite{Zhou18,Srinivasan19,Xu19} can be extremely effective even for complex real-world scenes.
However the assumption of multiple views severely limits their applicability, since the vast majority of images 
are not accompanied by views from other angles.

Another approach is to train a convolutional network to estimate depth from images~\cite{Eigen14,Li18}, enabling single-image view synthesis in realistic scenes~\cite{Niklaus19}.
Unfortunately this approach requires a training dataset of images with ground-truth depth.
Worse, depth predictors may not generalise beyond the scene types on which they are trained (\eg a network trained on indoor scenes will not work on outdoor images) so this approach can only perform view synthesis on scene types for which ground-truth depth can be obtained.

To overcome these shortcomings, there has been growing interest in view synthesis methods that do not use any 3D information during training.
Instead, an end-to-end generative model with 3D-aware intermediate representations can be trained from image supervision alone. Existing methods have shown promise on synthetic scenes of single objects~\cite{Kulkarni15,Tatarchenko16,Worrall17,Sitzmann19,Sitzmann19b}, but have been unable to scale to complex real-world scenes.
In particular, several recent methods represent 3D structure using dense voxel grids of latent features~\cite{Sitzmann19,Martin18}. With voxels, the fidelity of 3D information that can be represented is tied to the voxel dimensions, thus limiting the output resolution. On the other hand, point clouds are more flexible, generalise naturally to varying resolutions and are more efficient.

In this paper we introduce \modelname{}, a model for view \textbf{syn}thesis from a \textbf{sin}gle image in complex real-world scenes.
\modelname{} is an end-to-end model trained without any ground-truth 3D supervision.
It represents 3D scene structure using a high-resolution point cloud of learned features,
predicted from the input image using a pair of convolutional networks.
To generate new views from the point cloud, we render it from the target view using a high-performance differentiable point cloud renderer.
\modelname{} models scene semantics by building upon recent advances in generative models \cite{Brock19},
and training adversarially against learned discriminators.
Since all model components are differentiable, \modelname{} is trained end-to-end using image pairs and their relative camera poses;
at test-time it receives only a single image and a target viewpoint.

We evaluate our approach on three complex real-world datasets: Matterport~\cite{Matterport3D}, RealEstate10K~\cite{Zhou18}, and Replica~\cite{Replica19}.
All datasets include large angle changes and translations, increasing the difficulty of the task.
We demonstrate that our approach generates high-quality images and outperforms baseline methods that use voxel-based 3D representations.
We also show that our trained models can generalise at test-time to high-resolution output images, and even to new datasets with novel scene types.

\begin{figure*}[h]

\includegraphics[width=\textwidth]{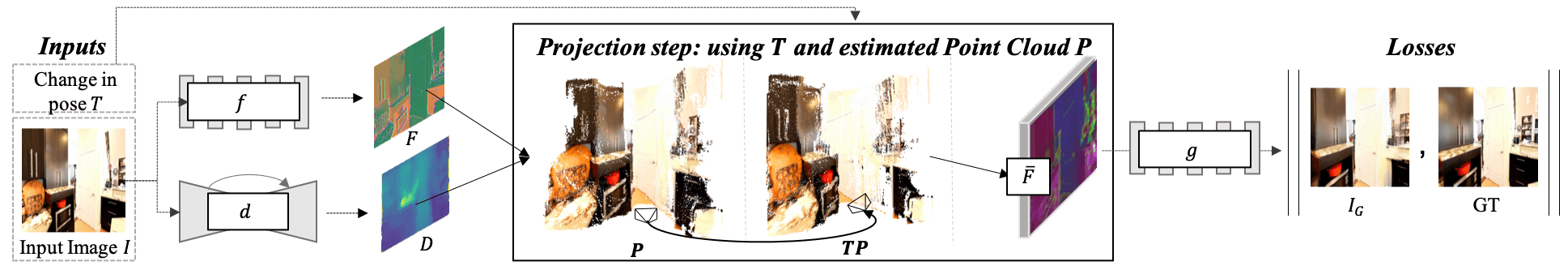}
\caption{{\bf Our end-to-end system. }
The system takes as input an image $I$ of a scene and change in pose $\m T$.
The {\em spatial feature predictor} ($f$) learns a set of features $F$ (visualised by projecting features using PCA to RGB) and the {\em depth regressor} ($d$) a depth map $D$. 
$F$ are projected into 3D (the diagram shows RGB for clarity) to give a point cloud $\mathcal{P}$ of features.
$\mathcal{P}$ is transformed according to $\m T$ and rendered.
The rendered features $\bar{F}$ are passed through the {\em refinement network} ($g$) to generate the final image $I_G$. 
$I_G$ should match the target image, which we enforce using a set of discriminators and photometric losses.}
\label{fig:overview}
\end{figure*}

\section{Related work}

Research into new view synthesis has a long history in computer vision.
These works differ based on whether they use multiple images or a single image at test time and on 
whether they require annotated 3D or semantic information.

\noindent {\bf View synthesis from multiple images.}
If multiple images of a scene can be obtained, inferred 3D geometry can be used to reconstruct the scene and then generate new views.
Traditionally, this was done using depth maps \cite{Penner17,Chaurasia13} or multi-view geometry \cite{Debevec96,Seitz06,Zitnick04,Fitzgibbon05,Kopf13,Debevec98}. 

In the learning era, DNNs can be used to learn depth.
\cite{Choi19,Hedman18,Meshry19,Martin18,Aliev19,Novotny19} use a DNN to improve view synthesis from a set of noisy, incomplete, or inconsistent depth maps. 
Given two or more images of a scene within a small baseline, \cite{Srinivasan17,Flynn19,Srinivasan19,Tulsiani18,Zhou18,Xu19} show impressive results at synthesising views within this narrow baseline.
\cite{Sitzmann19,Lombardi19,Olszewski19} learn an implicit voxel representation of one object given many training views and generate new views of that object at test time. 
\cite{Elsami18} use no implicit 3D representation.
Unlike these methods, we assume only one image at test time. 

\noindent {\bf View synthesis from a single image using ground-truth depth or semantics.}
A second vein of work assumes a large dataset of images with corresponding ground-truth 3D and semantic information to train their 3D representation \cite{Niklaus19,Tulsiani18b,Shin19}.
These methods are reliant on a large scale benchmark and corresponding annotation effort.
The depth may be obtained using a depth or lidar camera \cite{Geiger13,Silberman12,Knapitsch17} or SfM \cite{Li18};
however, this is time-consuming and challenging, especially for outdoor scenes, often necessitating the use of synthetic environments.
We aim to make predictions anywhere, \eg the wood scene in \figref{fig:systemcomparison}, and in realistic settings, {\em without} 3D information or semantic labels.

\noindent {\bf View synthesis from a single image.}
DNNs can be used to learn view synthesis in an end-to-end fashion.
One such line of work synthesises new views using purely image to image transformations
\cite{Tatarchenko16,Zhou16,Park17,Sun18,Kulkarni15,Chen16b}. 
Later work performs 3D operations directly on the learned embedding \cite{Worrall17} or interprets the latent space as an implicit surface \cite{Sitzmann19b}.
However, these works consider synthetic datasets with a single object per image and train one model per object class.
Most similar to ours is the recent work of \cite{Chen19}. 
However, they do not consider larger movements that lead to significant holes and dis-occlusions in the target image.
They also consider a more constrained setup; they consider synthetic object classes and mostly forward motion in KITTI \cite{Geiger13}, whereas we use a variety of indoor and outdoor scenes.

Many works explore using a DNN to predict 3D object shapes \cite{Groueix18,Gkioxari19,Tulsiani17,Yan16,Kanazawa18,Insafutdinov18} or the depth of a scene given an image \cite{Chen16,Eigen14,Li18,Zhou17}. These works focus on the quality of the 3D predictions as opposed to the view-synthesis task.

\noindent {\bf Generative models.}
We build on recent advances in generative models to produce high-quality images with DNNs~\cite{Goodfellow14a,Brock19,Karras19,Nguyen19,Park19}.
In \cite{Brock19,Karras19}, moving between the latent codes of different instances of an object class seemingly interpolates pose,
but explicitly modifying pose is hard to control and evaluate.
\cite{Nguyen19} allows for explicit pose control but not from a given image; they also use a voxel representation, which we find to be  computationally limiting.

\section{Method}
In this section, we introduce \modelname{} (\figref{fig:overview}) and in particular how we overcome
the two main challenges of representing 3D scene structure and scene semantics.
To represent the 3D scene structure, we project the image into a latent feature space which is in turn transformed using a differentiable point 
cloud renderer. This renderer injects a 3D prior into the network, 
as the predicted 3D structure must obey geometric principles.
To satisfy the scene semantics, we frame the entire end-to-end system as a GAN and build on architectural
innovations of recent state-of-the-art generative models.

\modelname{} takes an input image $I$ and relative pose $\m T$.
The input image is embedded to a feature space $F$ via a {\em spatial feature predictor} ($f$),  and a depth map $D$ via a {\em depth regressor} ($d$).
From $F$ and $D$, a point cloud $\mathcal{P}$ is created which is rendered into the new view ({\em neural point cloud renderer}).
The {\em refinement network} ($g$) refines the rendered features to give the final generated image $I_G$.
At training time, we enforce that $I_G$ should match the target image ({\em discriminator}).

\subsection{Spatial feature and depth networks}
Two networks, $f$ and $d$, are responsible for mapping the raw input image into a higher dimensional feature map and a depth map, respectively.
The spatial feature network predicts feature maps at the
same resolution as the original image.
These feature maps should represent scene semantics, \ie a higher-level representation than simply RGB colours.
The depth network estimates the 3D structure of the input image at the same resolution.
The depth does not have to be (nor would we expect it to be) perfectly accurate; however, it is explicitly learned in order to perform the task.
The design for $f$ and $d$ follows standard architectures built for the two tasks respectively: 

\noindent{\bf Spatial feature network $f$.} 
We build on the BigGAN architecture~\cite{Brock19} and use 8 ResNet blocks that maintain image resolution;
the final block predicts a $64$-dimensional feature for each pixel of the input image. 

\noindent{\bf Depth network $d$.}
We use a UNet \cite{Ronneberger15} with 8 downsampling and upsampling layers to give a final prediction of the same spatial resolution as the input.
This is followed by a sigmoid layer and a renormalisation step so the predicted depths fall within the per-dataset min and max values.
Please refer to the supplement for the precise details.

\begin{figure*}
\centering
\includegraphics[width=\linewidth]{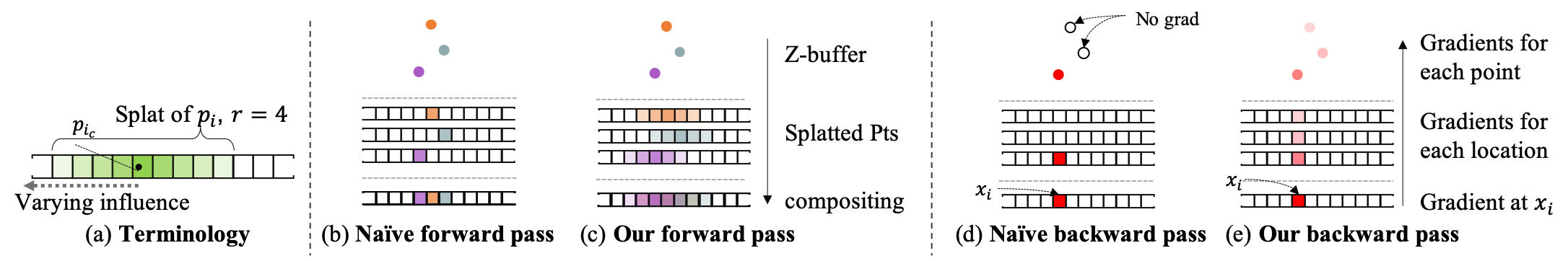}
\caption{Comparison of our rendering pipeline to a na\"{i}ve version. Given a set of points ordered in a z-buffer, our renderer projects points to a region of radius $r$ using $\alpha$-compositing, not just the nearest point.
When back-propagating through our renderer, gradients flow not just to the nearest point, but to all points in the z-buffer. (For simplicity we show 1D projections.) 
}
\label{fig:ptcld}
\end{figure*}

\subsection{Neural point cloud renderer}
\label{sec:3Drepresentation}
We combine the spatial features $F$ and predicted depths $D$ to give a 3D point cloud of feature vectors $\mathcal{P}$.
Given the input view transform $\m T$, we want to view this point cloud at the target viewpoint. 
This requires rendering the point cloud.
Renderers are used extensively in graphics, as reviewed in \cite{Kobbelt04,Sainz04},
but they usually focus on forward projection.
Our 3D renderer is a component of an end-to-end system, which is jointly optimised, and so needs to allow for gradient propagation; we want to train for depth prediction {\em without} any 3D supervision but only with a loss on the final rendered image.
Additionally, unlike traditional rendering pipelines, we are not rendering RGB colours but {\em features}.

\noindent{\bf Limitations of a na\"{i}ve renderer.} 
A na\"{i}ve renderer projects 3D points $p_i$ to one pixel or a small region -- the {\em footprint} -- in the new view.
Points are sorted in depth using a z-buffer.
For all points in the new view, the nearest point in depth (using the z-buffer) is chosen to colour that point.
A non-differentiable renderer does not provide gradients with respect to the point cloud positions (needed to train our depth predictor) nor the feature vectors (needed to train our spatial feature network).
Simply making the operations of a na\"{i}ve renderer differentiable is problematic for two reasons (illustrated in \figref{fig:ptcld}).
{\bf (1) Small neighbourhoods:} each point projects to only one or a few pixels in the rendered view. In this case, there are only a few gradients for each point in the $xy$-plane of the rendered view; this drawback
of {\em local} gradients is discussed in \cite{Jiang19} in the context of bilinear samplers.
{\bf (2) The hard z-buffer:} each rendered pixel is only affected by the nearest point in the z-buffer (\eg if a new pixel becomes
closer in depth, the output will suddenly change).

\noindent{\bf Our solution.}
We propose a neural point cloud renderer in order to solve the prior two problems by softening the hard decisions, as in \figref{fig:ptcld}.
This is inspired by \cite{Liu19}, which introduces a differentiable renderer for meshes by similarly softening the hard rasterisation decisions
and \cite{Zwicker01} which renderers point clouds by splatting points to a region and accumulating.
First, to solve the issue of small neighbourhoods, we splat 3D points to a disk of varying influence controlled by hyperparameters $r$ and $M$.
Second, to solve the issue of the hard z-buffer, we accumulate the effects of the $K$ nearest points, not just the nearest point, using a hyperparameter $\gamma$.

Our renderer first projects $\mathcal{P}$ onto a 2D grid under the given transformation $\m T$.
A 3D point $p_i$ is projected and splatted to a region with centre $p_{i_c}$ and radius $r$.
The influence of the 3D point $p_i$ on a pixel $l_{xy}$ is proportional
 to the Euclidean distance $d_2(\cdot,\cdot)$ from  the centre of the region:
 \begin{align*}
 \mathcal{N}(p_i, l_{xy}) &= 0 & \text{if} \quad d_2 (p_{i_c}, l_{xy}) > r \\
\mathcal{N}(p_i, l_{xy}) &= 1 - \frac{d_2 ( p_{i_c}, l_{xy})}{M} & \text{otherwise}.
 \end{align*}
 Though $\mathcal{N}$ is not differentiable, we can approximate derivatives using the subderivative.
 $r$ and $M$ control the spread and fall-off of the influence of a 3D point.

The projected points are then accumulated in a $z$-buffer; they are sorted according to their distance $d_i$ from the new camera and only the $K$ nearest points kept for each pixel in the new view.
The sorted points are accumulated using alpha over-compositing (where $\gamma$ is a hyperparameter):
\begin{align}
\label{eq:projection}
\rho_{i_{mn}} &= \mathcal{N}(p_i, l_{mn}) \\
\bar{F}_{mn} &= \sum_{i=1}^K \rho_{i_{mn}}^\gamma F_i \prod_{j=1}^{i-1} (1 - \rho_{j_{mn}}^\gamma),
\end{align}
where $\bar{F}$ is the projected feature map in the new view and $F$ in the original view.
$\gamma$ controls the blending; if $\gamma=0$, this is hard z-buffering.
This setup is illustrated in \figref{fig:ptcld}. 

\noindent{\bf Implementation.}
Our renderer must be high-performance, since we process batches of high-resolution point clouds during training.
We implement our renderer using a sequence of custom CUDA kernels, building upon work on high-performance triangle rasterisation with CUDA \cite{Laine11}.
We use a two-stage approach: in the first stage we break the output image into tiles, and determine the set of points whose footprint intersects each tile.
In the second stage, we determine the $K$ nearest points for each pixel in the output image, sorting points in depth using per-pixel priority queues in shared memory to reduce global memory traffic.

\noindent{\bf Other approaches.} This method is related to the point cloud rasterisers of \cite{Yifan19,Insafutdinov18,Aliev19}.
However, our renderer is a simpler than \cite{Yifan19} and we apply it in an end-to-end framework. 
While \cite{Aliev19} also renders point clouds of features, they only back-propagate to the feature vectors, not the 3D positions.
\cite{Insafutdinov18} stores the predicted points in a voxel grid before performing the projection step; this limits the resolution. 

\noindent{\bf Performance.}
On a single V100 GPU, rendering a batch of six point clouds with $512^2=262{,}144$ points each to a batch of six images of size $256\times256$ takes 36ms for the forward pass and 5ms for the backward pass.
In contrast, converting the same point cloud to a $256^3$ voxel grid using the implementation from~\cite{Insafutdinov18} takes $\approx1000$ms for the forward pass and $\approx2000$ms for the backward pass.

\begin{figure*}[t]
\centering

\label{fig:realestate10k}
\includegraphics[width=\textwidth]{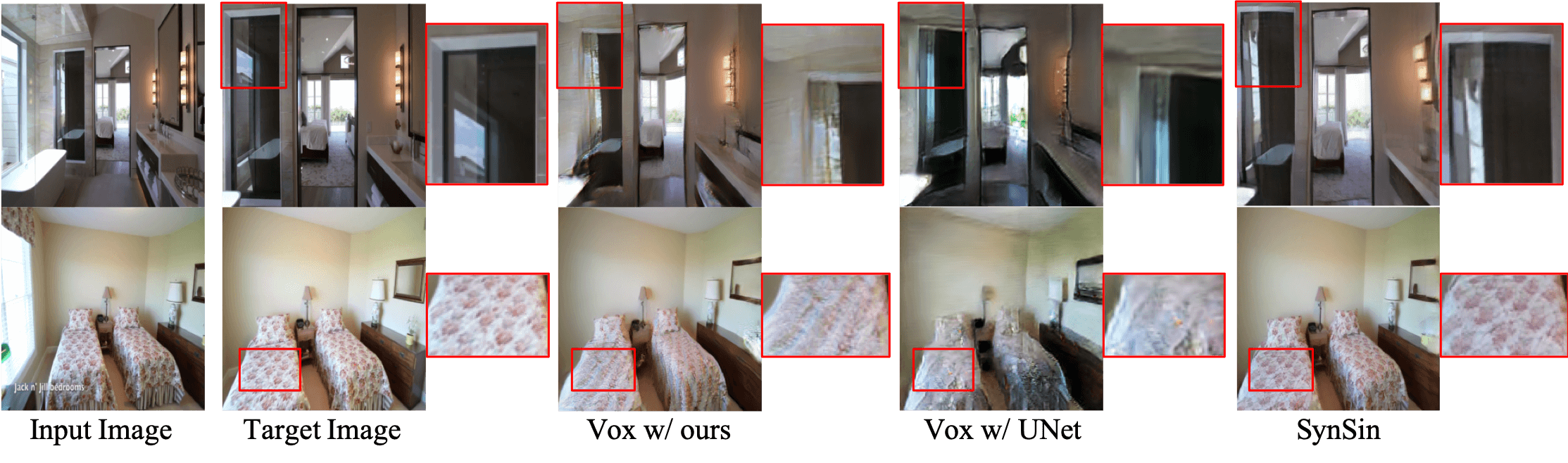}

%
  \caption{Qualitative results on RealEstate for ours and baseline methods. Given the input view and the camera parameters, the methods are tasked to produce the target image. 
  The red squares denote interesting differences between the methods. In the upper row, our model better recreates the true 3D; in the bottom row, our model is better able to preserve detail.}
  \label{fig:qualresults}
  \vspace{-0.5em}
\end{figure*}

\subsection{Refinement module and discriminator}
\label{sec:architecture}
Even if the features are projected accurately, regions not visible in the input view will be empty in the target view.
The refinement module should inpaint~\cite{Bertalmio00,Criminisi04} these missing regions
in a semantically meaningful (\eg missing portions of a couch should be filled in with similar texture) and
geometrically accurate (\eg straight lines should continue to be straight) manner.
To solve this task, we take inspiration from recent generative models \cite{Brock19,Karras19,Park19}.

Deep networks have been previously applied to inpainting~\cite{Pathak16,Teterwak19,Wang19}.
In a typical inpainting setup, we know a-priori which pixels are correct and which need to be synthesised.
In our case, the refinement network should perform two tasks.
First, it should inpaint regions with no projected features, \eg regions on the image boundary or dis-occluded regions.
The refinement module can discover these regions, as the features have values near zero.
Second, the refinement module should correct local errors (\eg noisy regions resulting from noisy depth).

To build the refinement network, we use 8 ResNet \cite{He16} blocks, taking inspiration
from \cite{Brock19}.
Unlike \cite{Brock19}, we aim to generate a new image conditioned on an input view not a random vector.
Consequently, we find that it is important to maintain the image resolution as much as possible to obtain high quality results.
We modify their ResNet block to create a downsampling block.
The downsampling block is used to decrease the image resolution by two sizes before upsampling to the original image resolution.
To model the ambiguity in the inpainting task, we use batch normalisation injected with noise \cite{Brock19}.
We additionally apply spectral normalisation following each convolutional layer \cite{Zhang18}.

The GAN architecture and objective used is that of \cite{Wang18}.
We use 2 multi-layer discriminators at a lower and higher resolution and a feature matching loss on the discriminator.

\subsection{Training}
\noindent{\bf Training objective.}
\label{sec:losses}
The network is trained with an L1 loss, content loss and discriminator loss between the generated and target image.
The total loss is then 
$\mathcal{L} = \lambda_{GAN} \mathcal{L}_{GAN}  + \lambda_{l1} \mathcal{L}_{l1} + \lambda_{c} \mathcal{L}_{c}$.

\noindent{\bf Training details.}
The models are trained with the Adam optimiser using a 0.01 learning rate for the discriminator, 0.0001 for the
generator and momentum parameters (0, 0.9).
$\lambda_{GAN} = 1$, $\lambda_c = 10$, $\lambda_{l1}$ = 1.
$\gamma = 1$, $r = 4$ pixels, $K=128$, $W=H=256$.
The models are trained for $50$K iterations.
We implement our models in PyTorch \cite{Paszke17}; they take 1-2 days to train on 3 Tesla V100 GPUs.

\begin{table*}
\centering
\begin{minipage}{\linewidth}

\resizebox{\textwidth}{!}{
\begin{tabular}{lccccccccccccccc}
\toprule
    &\multicolumn{9}{c}{\bf Matterport \cite{Matterport3D}} & \multicolumn{3}{c}{\bf RealEstate10K  \cite{Zhou18}} & \multicolumn{3}{c}{\bf Replica \cite{Replica19}} \\
    \cmidrule(lr){2-10} \cmidrule(lr){11-13} \cmidrule(lr){14-16}
    & \multicolumn{3}{c}{PSNR $\uparrow$} & \multicolumn{3}{c}{SSIM $\uparrow$} & \multicolumn{3}{c}{Perc Sim $\downarrow$} & PSNR $\uparrow$ & SSIM $\uparrow$ & Perc Sim $\downarrow$  & PSNR $\uparrow$ & SSIM $\uparrow$ & Perc Sim $\downarrow$ \\ 
 		& Both & InVis & Vis & Both & InVis & Vis & Both & InVis & Vis  & \\ 
\cmidrule(lr){2-4} \cmidrule(lr){5-7} \cmidrule(lr){8-10} \cmidrule(lr){11-13} \cmidrule(lr){14-16}
1.\ \modelname{} (small ft) & {\bf 21.14}  & {\bf 20.19}  & {\bf 21.84} & {\bf 0.71}  & 0.70  & 0.69  & {\bf 1.68} & 0.45 & {\bf 0.98} & 21.10$_{3.48}$ & 0.73$_{0.14}$ & 1.34$_{0.55}$ & {\bf 22.36} & 0.80 & 1.64   \\
2.\ \modelname{} (hard z) & 21.08 & 20.23 & 21.70 & 0.70 & 0.70 & 0.67 & 1.82 & 0.44 & 1.11 & 21.40$_{4.06}$ & 0.70$_{0.15}$ & 1.45$_{0.61}$ & 20.70 & 0.76 & 1.95 \\
3.\ \modelname{} (rgb) & 20.64  & 19.87  & 21.21  & 0.67  & 0.69 & 0.65 & 2.06 & 0.49 & 1.27 & 20.92$_{3.81}$ & 0.68$_{0.14}$ & 1.67$_{0.51}$ & 20.44 & {0.75} & 2.03  \\
4.\ \modelname{} & {20.91} & {19.80} & {21.62} & {\bf 0.71} & {\bf 0.71} & {\bf 0.70} & {\bf 1.68} & {\bf 0.43} & {\bf 0.99}  & {\bf 22.31$_{4.97}$} & {\bf 0.74$_{0.16}$} & {\bf 1.18$_{0.64}$} & {21.94} & {\bf 0.81} & {\bf 1.55} \\
    \midrule \midrule
5.\ \modelname{} (w/ GT) & 22.65 & 19.64 & 26.19 & 0.78 & 0.71 & 0.82 & 1.37 & 0.50 & 0.64 & -- & -- & -- & 23.72 & 0.86 & 1.22 \\
6.\ \modelname{} (sup.\ by GT) & 21.59 & 20.32 & 22.46 & 0.72 & 0.71 & 0.71 & 1.60 & 0.43 & 0.92 & -- & -- & -- & 22.54 & 0.80 & 1.55 \\
    \midrule
7.\ Im2Im 					    & 15.87 & 16.20 & 15.97 & 0.53 & 0.60 & 0.48 & 2.99 & 0.58 & 2.05 & 17.05$_{4.78}$ & 0.56$_{0.18}$ & 2.19$_{1.22}$ & 17.42 & 0.66 & 2.29  \\
8.\  Vox w/ UNet 				    & 18.52  & 17.85 & 19.05 & 0.57 & 0.57 & 0.57 & 2.98 & 0.77  & 1.96 & 17.31$_{2.63}$ & 0.53$_{0.15}$ & 2.30$_{0.40}$ & 18.69 & 0.71 & 2.68 \\
9.\ Vox w/ ours 				    & 20.62 & 19.64 & 21.22 & 0.70 & 0.69 & 0.68 & 1.97 & 0.47 & 1.19 & 21.88$_{4.39}$ & 0.71$_{0.15}$ & 1.30$_{0.55}$ & 19.77 & 0.75 & 2.24 \\

\bottomrule

\end{tabular}
}
\caption{Results on Matterport3D \cite{Matterport3D}, RealEstate10K \cite{Zhou18}, and Replica \cite{Replica19}.  $\uparrow$ denotes higher is better, $\downarrow$ lower is better.
$XX_{YY}$ denotes std dev.\ $YY$.
The ablations demonstrate the utility of each aspect of our model. We outperform all baselines for both datasets and are nearly as good as a model supervised with depth (\modelname{} (sup.\ by GT)).
We also perform best when considering regions visible (Vis) and not visible (InVis) in the input view.}
\label{tab:habitatresults}
\vspace{-1em}
\end{minipage}
\end{table*}

\section{Experiments}
We evaluate our approach on the task of view synthesis using novel real-word scenes.
We validate our design choices in~\secref{sec:comparisons} by ablating our approach and comparing against competing 
end-to-end view synthesis pipelines.
We also compare to other systems and find that our model performs better than one
based on a trained depth predictor, which fails to generalise well to the new domain.
We additionally evaluate \modelname{}'s generalisation performance to novel domains (\secref{sec:generalreplica}) as well as higher image resolutions (\secref{sec:higher}). 
Finally, we use \modelname{} to synthesise trajectories from an initial image in \secref{sec:synthesis}, demonstrating that it
can be used for a walk-through application.
Additional results are given in the supplement.

\subsection{Experimental setup}

\noindent{\bf Datasets.}
We focus on using realistic data of indoor and outdoor environments as opposed to synthetic objects.

The first framework we use is Habitat \cite{Habitat19}, which allows for testing in a variety of scanned indoor scenes.
The Habitat framework can efficiently generate image and viewpoint pairs for an input scene. 
We use two sources of indoor scenes: Matterport3D \cite{Matterport3D}, consisting of reconstructions of homes, and Replica \cite{Replica19}, which consists of higher fidelity scans of indoor scenes. 
The Matterport3D dataset is divided at the scene level into train/val/test which contain 61/11/18 scenes.
The Replica dataset is only used at evaluation time to test generalisability.
Pairs of images are generated by randomly selecting a viewpoint in a scene and then randomly modifying the
viewing angle in a range of $\pm 20^{\circ}$ in each Euclidean direction and the position within $\pm 0.32$m.

The second dataset we use is RealEstate10K \cite{Zhou18}, which consists of videos of walkthroughs of properties and the corresponding camera parameters (intrinsic and extrinsic)
obtained using SfM.
The dataset contains both indoor and outdoor scenes.
It is pre split into a disjoint set of train and test scenes; we subdivide train into a training and validation set to give approximately 57K/14K/7K scenes in train/val/test.
The scenes in the test set are  unseen. 
We sample viewpoints by selecting a reference video frame and then selecting a second video frame a maximum of 30 frames apart.
In order to sample more challenging frames, we choose pairs with a change in angle of  $> 5^{\circ}$ and a change in
position of greater than $0.15$ if possible (see \cite{Zhou18} for a discussion on metric scale). 
To report results, we randomly generate a set of 2000 pairs of images from the test set.

\noindent{\bf Metrics.} 
Determining the similarity of images in a manner correlated with human judgement is challenging \cite{Zhang18b}.
We report multiple metrics to obtain a more robust estimate of the relative
quality of images.
We report the PSNR, SSIM, and perceptual similarity of the images generated by the different models.
Perceptual similarity has been recently demonstrated
to be an effective method for comparing the similarity of images \cite{Zhang18b}.
Finally, we validate that these metrics do indeed correlate with human judgement by performing a user study on Amazon
Mechanical Turk (AMT).

\subsection{Baselines}
We first abate the need for a soft differentiable renderer by comparing to variants with a small footprint, hard z-buffering, and that directly project RGB values.
These models use the same setup, training schedule, and sequence of input images/viewpoints as \modelname{}.

\noindent {\bf \modelname{} (small ft): }
We set $K=128$ and $r=0.5$ in our model to investigate the utility of a large footprint.

\noindent{\bf  \modelname{} (hard z): }
We set $K=1$ and $r=4$ in our model to investigate the utility of the soft z-buffer.

\noindent{\bf  \modelname{} (rgb): } We project RGB values not features.

\modelname{} does not assume ground-truth depth at test time; the depth predictor is trained end-to-end for the given task.
We investigate the impact of ground-truth (GT) depth by reporting two variants of our model. These models act as upper bounds and can only be trained on Matterport3D (not RealEstate10K), as they use true depth information.

\noindent{\bf \modelname{} (w/ GT):} The true depth is used as $D$.

\noindent{\bf \modelname{} (sup.\ by GT):} $D$ is supervised by the true depth. (In all other cases \modelname{}'s $D$ is learned with {\em no} supervision).

We evaluate our 3D representation by comparing to a method that uses no 3D and one that uses voxels.
 As no
methods exist for the challenging datasets we consider, we re-implement the baselines for a fair comparison.
The baselines use the same setup, training schedule, and sequence of input images/viewpoints as \modelname{}.

\noindent{\bf Im2Im:} This baseline evaluates an image-to-image method; we re-implement \cite{Zhou16}.
\cite{Zhou16} only considered a set of discretised rotations about the azimuth and a smaller set of rotations in elevation.
However, the changes in viewpoint in our datasets arise from rotating continuously in any direction and translating in 3D.
We modify their method to allow for these more complex transformations.

\noindent {\bf Vox:} 
This baseline swaps our implicit 3D representation for a voxel based representation.
The model is based on that of \cite{Sitzmann19}. 
However, \cite{Sitzmann19} trains one model per object, so their model effectively learns to interpolate between the $>$100 training views  unlike our model,
which extrapolates to new real-world test scenes given a {\em single} input view.
We consider two variants: {\bf Vox w/ UNet} uses the UNet encoder/decoder of \cite{Sitzmann19} whereas {\bf Vox w/ ours} uses a similar ResNet encoder/decoder setup to \modelname{}.
This comparison evaluates our 3D approach as opposed to a voxel based one as well as whether our encoder/decoder setup is preferable.

Finally, we compare \modelname{} to existing pipelines that perform view synthesis. These systems make different assumptions and follow different approaches.
This comparison validates our use of a learned end-to-end system.

\noindent {\bf  StereoMag \cite{Zhou18}:}
This system takes two images as input at test time.
Assuming two input views simplifies the problem of 3D understanding compared to our work, which estimates 3D from a single view.

\noindent {\bf 3D View:}
This system trains a single-image depth predictor on images with ground-truth depth (\eg MegaDepth~\cite{Li18}).
Predicted depths are used to convert the input image to a textured 3D mesh, which is extended in space near occlusion boundaries using isotropic colour diffusion \cite{Hedman18b}.
Finally the mesh is rendered from the target view.
The approach is similar to 3D Photos \cite{Photos3D}.

\begin{table*}
\begin{minipage}{\linewidth}
\centering
\begin{minipage}{0.33\linewidth}
\centering
\resizebox{\textwidth}{!}{
\begin{tabular}{lccc}
\toprule
    \multicolumn{4}{c}{\bf System comparison on RealEstate10K \cite{Zhou18}} \\
\cmidrule{1-4}
   &  PSNR $\uparrow$ & SSIM $\uparrow$ & Perc Sim $\downarrow$ \\ 
\cmidrule{2-4}
\modelname{} & 22.31$_{4.97}$ &  0.74$_{0.16}$ &  1.18$_{0.64}$  \\
\cmidrule{2-4}
3DView 	&  $21.88_{8.43}$ & $0.66_{0.22}$ & $1.52_{1.03}$  \\
StereoMag \cite{Zhou18} 	& $25.34_{9.48}$ & $0.82_{0.13}$ & $1.19_{0.77}$ \\
\bottomrule
\end{tabular}
}
\centering
  \caption{ \modelname{} performs better than a system trained with GT depth (3DView)
  and approaches the performance of \cite{Zhou18}, which uses 2 input views at test time.
  \label{tab:systemcomparison}
  }

\end{minipage}   \quad
\begin{minipage}{0.31\linewidth} 
\centering
\resizebox{\textwidth}{!}{
\begin{tabular}{lccc}
\toprule
    &\multicolumn{3}{c}{\bf Generalisation to higher res.} \\
    \cmidrule(lr){2-4}
 		& PSNR $\uparrow$ & SSIM $\uparrow$ & Perc Sim $\downarrow$ \\ 
\cmidrule(lr){2-4} 
\modelname{} & {\bf 22.06$_{6.30}$} & {\bf 0.72$_{0.18}$} & {\bf 1.00$_{0.65}$} \\

Vox w/ ours & 18.82$_{2.52}$ & 0.61$_{0.14}$ & 2.47$_{0.36}$ \\
\bottomrule
\end{tabular}
}
\caption{Results when applying models trained on $256\times256$ images
to $512\times512$ images.}
\label{tab:higherres}
\end{minipage} \quad 
\begin{minipage}{0.28\linewidth}
\centering
\vspace{3mm}
\resizebox{\textwidth}{!}{
\begin{tabular}{lccc}
\toprule
  \multicolumn{4}{c}{\bf AMT User Study} \\
\midrule
 		& Ours & Vox w/ ours & Neither \\ 
\cmidrule{2-4}
{\em E-O} & {\bf 68.7} & 31.3 & --   \\
{\em E-O-N } &  {\bf 55.6} & 27.3 & 17.2 \\
\bottomrule
\end{tabular}
}
\caption{
\% of videos chosen as most realistic.
 In {\em E-O}, users choose the better method; in {\em E-O-N}, users can say neither is better.
}
\label{tab:userstudy}
\end{minipage} 
\end{minipage}
\vspace{-0.5em}
\end{table*}

\begin{figure}

\includegraphics[width=\linewidth]{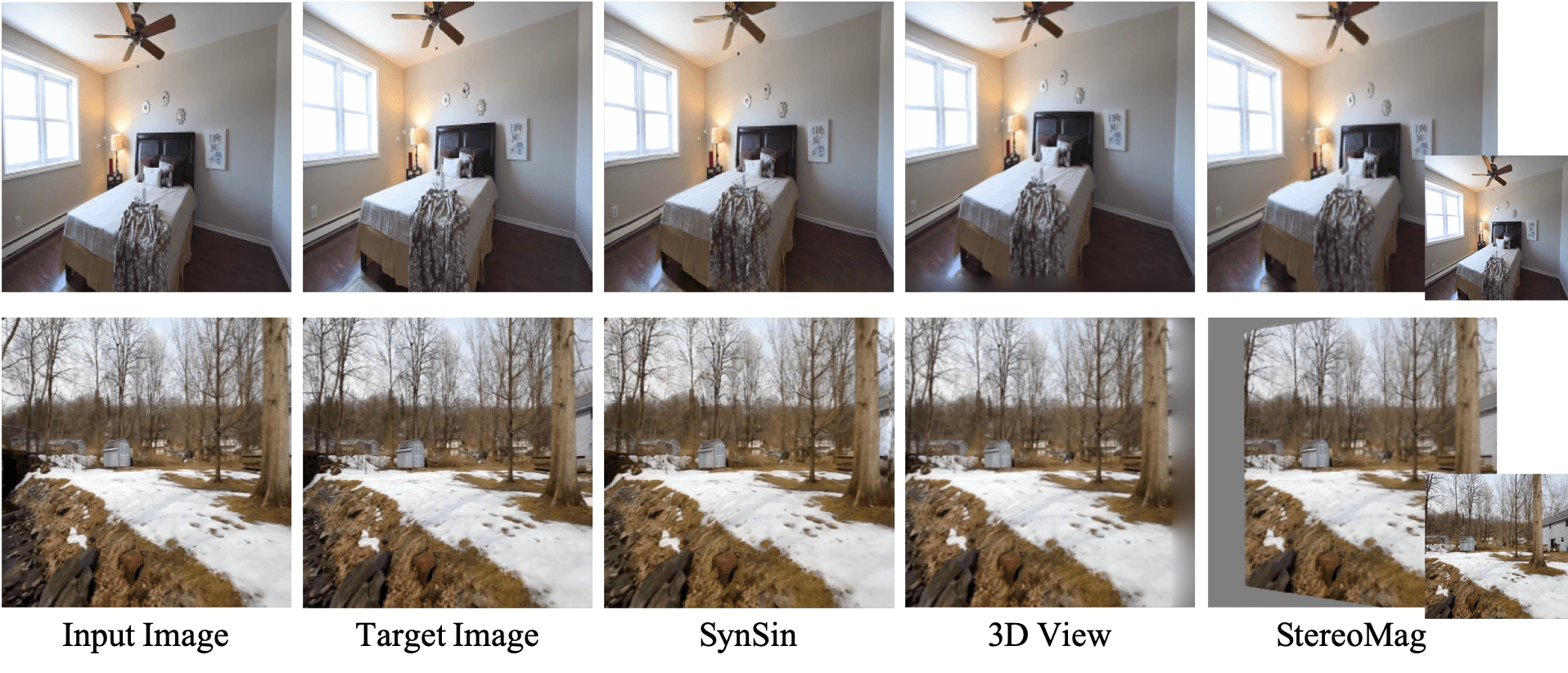}

\caption{System comparisons on RealEstate10K, illustrating failure cases. 
Note StereoMag~\cite{Zhou18} uses two input images (second is shown as an inset).
Unlike \cite{Zhou18} we inpaint missing regions (bottom row); \cite{Zhou18} fails to model the left region and cannot inpaint the missing region.
 3DView uses a model pretrained for depth, causing their system to produce inaccurate results in some cases (\eg the bed in the top row).}
\label{fig:systemcomparison}
\end{figure}

\subsection{Comparisons with other methods}
\label{sec:comparisons}
\noindent{\bf Results on Matterport3D and RealEstate10K.}
We train our models, ablations, and baselines on these datasets.

To better analyse the results, we  compare models on how well they understand the 3D scene structure and the scene semantics (discussed in \secref{sec:intro}).
To achieve this, we report metrics on the final prediction (Both) but also on the regions of the target image that are visible (Vis) and not visible (InVis) in the input image.
(Vis) evaluates the quality of the learned 3D scene structure, as it can be largely solved by accurate depth prediction.
(InVis) evaluates the quality of a model's understanding of scene semantics; it requires a holistic understanding of semantic and geometric properties to reasonably in-paint missing regions.
In order to determine the (Vis) and (InVis) regions, we use the GT depth in the input view to obtain a binary mask of which pixels are visible in the target image.
This is only possible on Matterport3D (RealEstate10K does not have GT depth).

\tabref{tab:habitatresults} and \figref{fig:qualresults} report results on Matterport3D and RealEstate10K.
On both datasets, we perform better than the baselines on all metrics and under all conditions, demonstrating the utility of both our 3D representation and our inpainting module.
These results demonstrate that the differentiable renderer is important for training the depth model (rows 1-4).
Our encoder decoder setup is shown to be important, as it improves the baseline's performance significantly (rows 8-9).
Qualitatively, our model preserves fine detail and predicts 3D structure better than the baselines.

\noindent {\bf System comparison on RealEstate10K.}
We compare our system to the 3DView and StereoMag \cite{Zhou18} in  \tabref{tab:systemcomparison} and \figref{fig:systemcomparison}.
Our model performs better than 3DView despite their method having 
been trained with hundreds of thousands of depth images.
We hypothesise that this gap in performance is due to the 3DView's depth prediction not generalising well; their dataset consists of images of mostly close ups of objects
whereas ours consists of scenes taken inside or outdoors.
This baseline demonstrates that using an explicit 3D representation is problematic when the test domain differs from the training domain,
as the depth predictor cannot generalise.
Finally, our method of inpainting is better than that of 3DView, which produces a blurry result.
\cite{Zhou18} does not inpaint unseen regions in the generated image.

\noindent{\bf Comparison with upper bounds.}
We compare our model to \modelname{} (w/ GT) and \modelname{} (sup.\ with GT) in \tabref{tab:habitatresults}.
These models either use GT depth or are supervised by GT depth; they are upper bounds of performance.
While there is a performance gap between \modelname{} and \modelname{} (w/ GT) under the (Vis) condition, this gap shrinks for the (InVis) condition.
Interestingly, \modelname{} trained with {\em no} depth supervision performs nearly as well as \modelname{} (sup.\ with GT) under both the (Vis) and (InVis) conditions; 
our model also generalises better to the Replica dataset.
This experiment demonstrates that having true depth during training does not necessarily give a large boost in a downstream task and could hurt generalisation performance.
It validates our decision to use an end-to-end system (as opposed to using depth estimated from a self-supervised method).

\begin{figure}
\centering
\includegraphics[width=\linewidth]{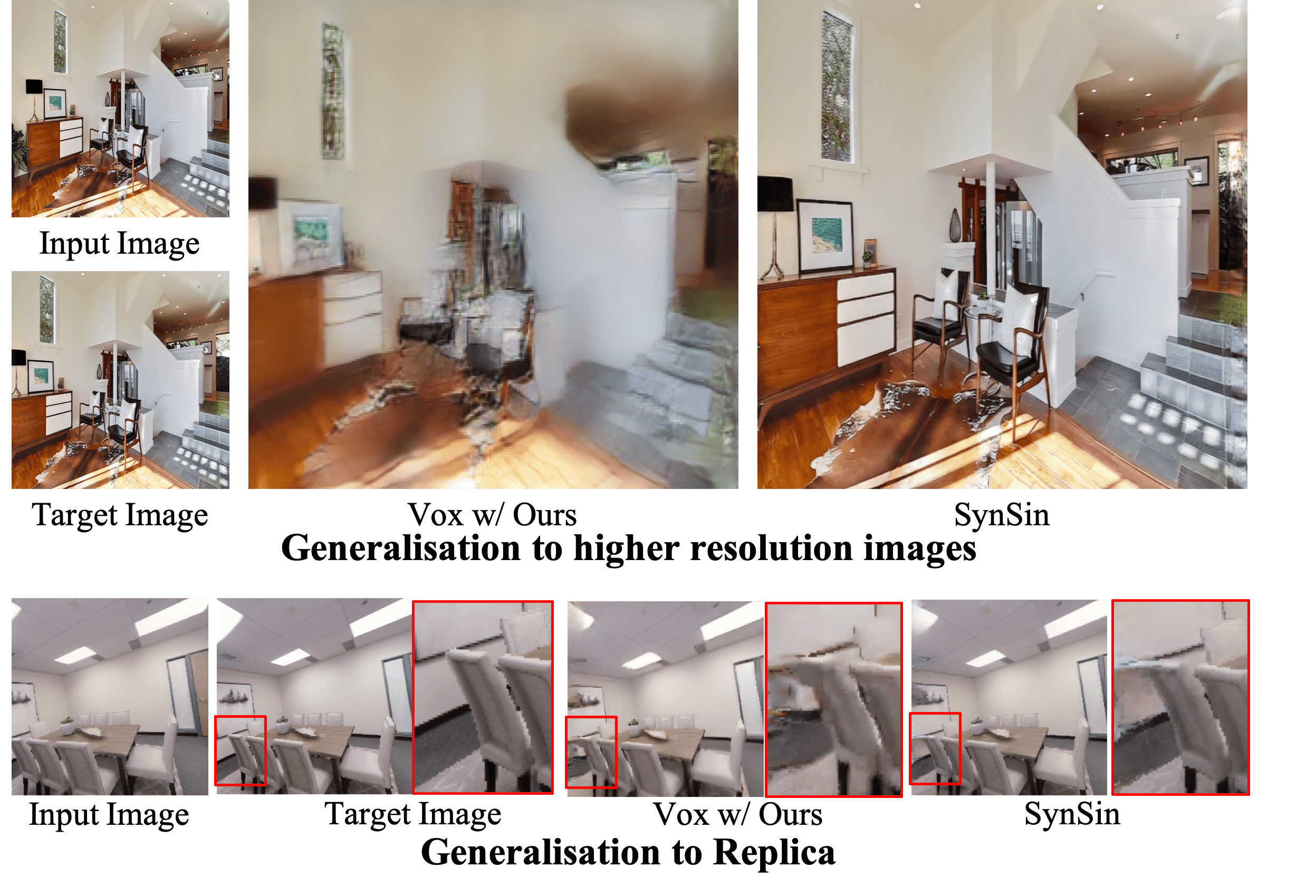}
\caption{Comparison of \modelname{} against the baseline, Vox w/ ours, at generalising to higher res $512 \times 512$ images and Replica 
\cite{Replica19}. Ours generalises better with fewer artefacts.}
\label{fig:higherres}
\end{figure}

\noindent{\bf Generalisation to Replica.}
\label{sec:generalreplica}
Given the models trained on Matterport3D, we evaluate generalisation performance (with no further fine-tuning) on Replica in \tabref{tab:habitatresults}.
Replica contains additional types of rooms (\eg office and hotel rooms) and is higher quality than Matterport (it has fewer geometric and lighting artefacts and more complex textures).
\modelname{} generalises better to this unseen dataset; qualitatively, \modelname{} seems to introduce fewer artefacts (\figref{fig:higherres}).

\subsection{Generalisation to higher resolution images}
\label{sec:higher}
We also evaluate generalisation to higher image resolutions in \tabref{tab:higherres} and \figref{fig:higherres}.
\modelname{} can be applied to higher resolution images without any further training.
The ability to generalise to higher resolutions is due to the flexible 3D representation in our approach:
 the networks are fully convolutional and the 3D point cloud can be sampled at any resolution to maintain the resolution of the features.
As a result, it is straightforward at test time to apply a network trained on a smaller image size (\eg $256 \times 256$) to one of a different size (\eg $512 \times 512$).
Unlike our approach, the voxel baseline suffers a dramatic performance drop when applied to a higher resolution image. 
This drop in performance is presumably a result of the heavy downsampling and imprecision resulting from
representing the world as a coarse voxel grid.

\subsection{Depth predictions}
\label{sec:depthprediction}
\begin{figure}
\centering

\includegraphics[width=\linewidth]{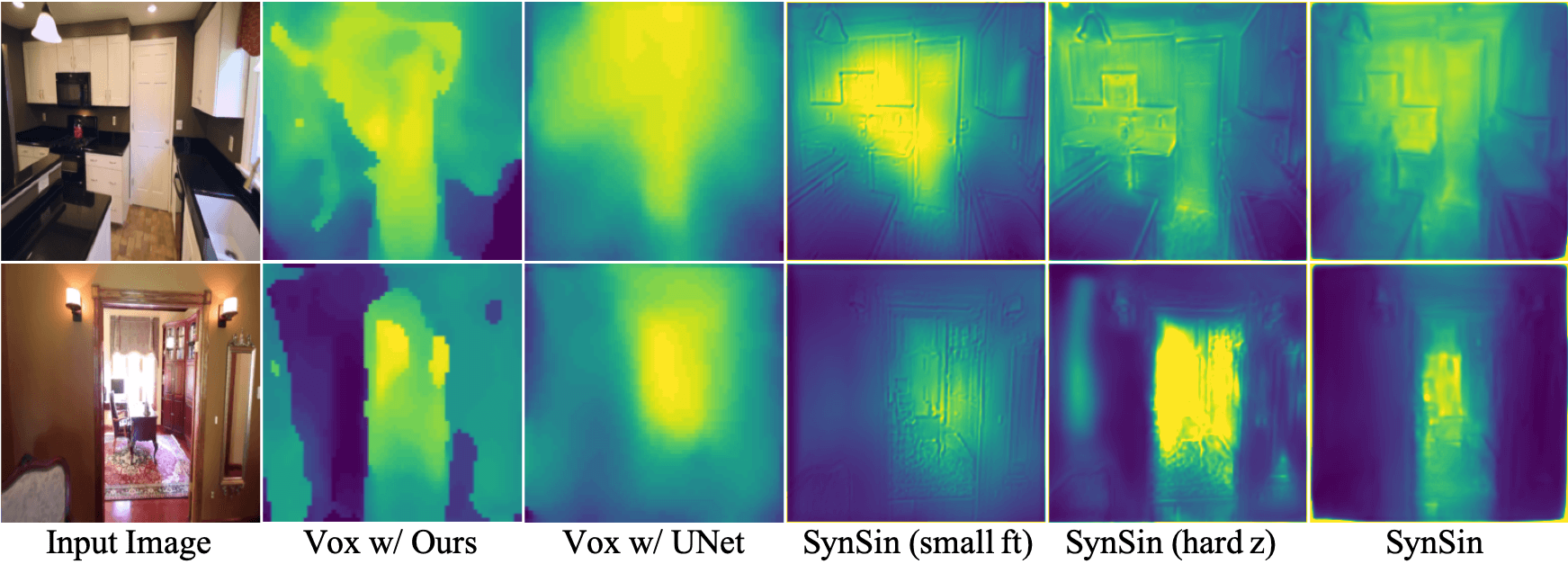}
\caption{Recovered depth predictions for both our method and the baselines.
The baselines predict a less accurate and coarser depth. Using a smaller radius or hard z-buffer
produces qualitatively similar or worse depth maps.}
\label{fig:ptclds3d}
\vspace{-0.5em}
\end{figure}

We evaluate the quality of the learned 3D representation qualitatively in \figref{fig:ptclds3d} for \modelname{} trained on RealEstate10K. 
We note that the accuracy of the depth prediction only matters in so far as it improves results on the view synthesis task.
However, we hypothesise that the quality
of the generated images and predicted depth maps are correlated, so looking at the quality of the depth maps should give some insight into the quality of the learned models.
The depth map predicted by our method is higher resolution and more realistic than the depth map predicted by the baseline methods.
Additionally, our differentiable point cloud renderer appears to improve the depth quality over using a hard z-buffer or a smaller footprint.
However, we note that small objects and finer details are not accurately recreated. This is probably because these structures have a limited impact
on the generated images.

\subsection{User study: Animating still images}
\label{sec:synthesis}
Finally, we task \modelname{} to synthesise images along a trajectory.
Given an initial input frame from a video in RealEstate10K,  \modelname{}
generates images at the camera position of the 30 subsequent frames.
While changes are hard to see in a figure (\eg \figref{fig:teaser}), the supplementary videos clearly show smooth motion 
and 3D effects.
These demonstrate that \modelname{} can generate reasonable videos despite being trained purely on images.
To evaluate the quality of the generated videos, we 
perform an AMT user study.

We randomly choose 100 trajectories and generate videos using \modelname{} and the Vox w/ ours baseline.
Five users are asked to rate which method's video is most realistic.
For each video, we take the majority vote to determine the best video.
We report the percentage of times the users choose a given method in \tabref{tab:userstudy}.

\noindent{\bf Either-or setup (E-O):}
Users rate whether the baseline or our generated video is more realistic.

\noindent{\bf Either-or-neither setup (E-O-N):}
Users rate whether the baseline or our generated video is more realistic or whether they are equally realistic/unrealistic ({\em neither}).
When taking the majority vote, if their is no majority, {\em neither} video is said to be more / less realistic

In both cases, users prefer our method, presumably because our videos have smoother motion
and fewer artefacts. 

\section{Conclusion}

We introduced \modelname{}, an end-to-end model for performing single image view synthesis.
At the heart of our system are two key components:
{\em first} a differentiable neural point cloud renderer, and
{\em second} a generative refinement module.
We verified that our approach can be learned end-to-end on multiple realistic datasets, generalises to unseen scenes,  
can be applied directly to higher image resolutions, and can be used to generate reasonable videos along a given trajectory.
While we have introduced \modelname{} in the context of view synthesis, we note that using a neural point cloud renderer
within a generative model
has applications in other tasks.

\noindent{\bf Acknowledgements}
The authors thank Johannes Kopf for sharing code, Manolis Savva and Erik Wijmans for help with the Habitat dataset, and Sebastien Ehrhardt, Oliver Groth, and Weidi Xie for feedback on paper drafts.

{\small
\bibliographystyle{ieee_fullname}
\bibliography{shortstrings,egbib,vgg_other,vgg_local}
}

\clearpage

\appendix

\twocolumn[{%
	\renewcommand\twocolumn[1][]{#1}%
	\begin{center}
	\bf \Large \modelname{}: Appendix
	\vspace{2em}
	\end{center}
	
}]

We give additional results in \secref{sec:qualresults},
additional architectural details in \secref{sec:archresults}, 
additional information about baselines in \secref{sec:baselines},
and finally information about datasets in \secref{sec:datasets}.
Finally, we discuss some choices that did and did not work in \secref{sec:negativeresults}.

\section{Additional experimental results}
\label{sec:qualresults}
\begin{figure}
\centering
\begin{overpic}[width=0.76\linewidth]{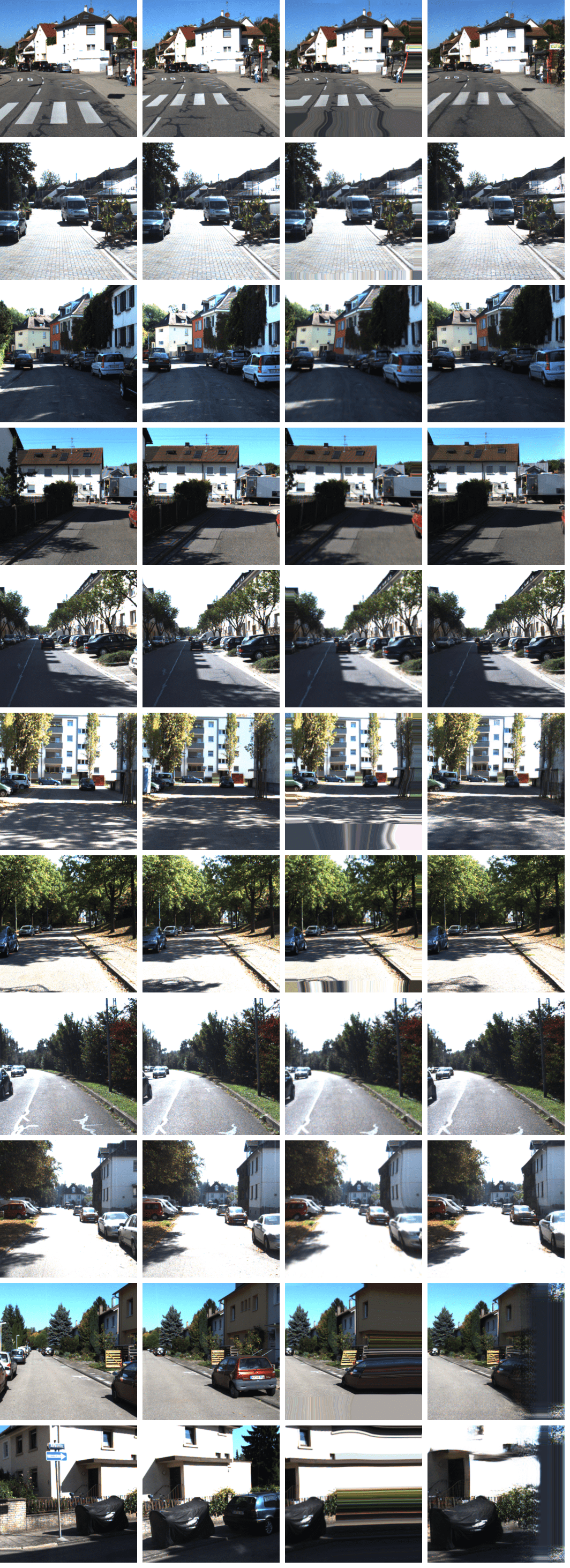}
\scriptsize
\put (2,-1.5) {Input Img}
\put (11,-1.5) {Target Img}
\put (19,-1.5) {ContView \cite{Chen19}}
\put (30,-1.5) {\modelname{}}
\end{overpic}
\vspace{2em}
\caption{Qualitative results on KITTI \cite{Zhou18} comparing \modelname{} to \cite{Chen19}.
The bottom two rows demonstrate failure cases due to large viewpoint change. Zoom in for details.}
\label{fig:kitti}
\end{figure}

\noindent {\bf Results on KITTI \cite{Geiger13}.}
We evaluated our model on the KITTI dataset in order to compare with \cite{Chen19}.
We trained our model on the KITTI dataset and compared with their pretrained model on a held-out set in \tabref{tab:kitti}.
We achieve similar or better results on all metrics.
Additionally, because \cite{Chen19} resamples the input image, it cannot generate pixels unseen in the original view.
As shown in \figref{fig:kitti} (rows 1,2,6), this causes severe artefacts for backward motion.

We additionally show some failure cases for both methods when the viewpoint change is much larger than the average viewpoint
change seen at test time in the bottom two rows.  

\begin{figure}
\begin{overpic}[width=\linewidth]{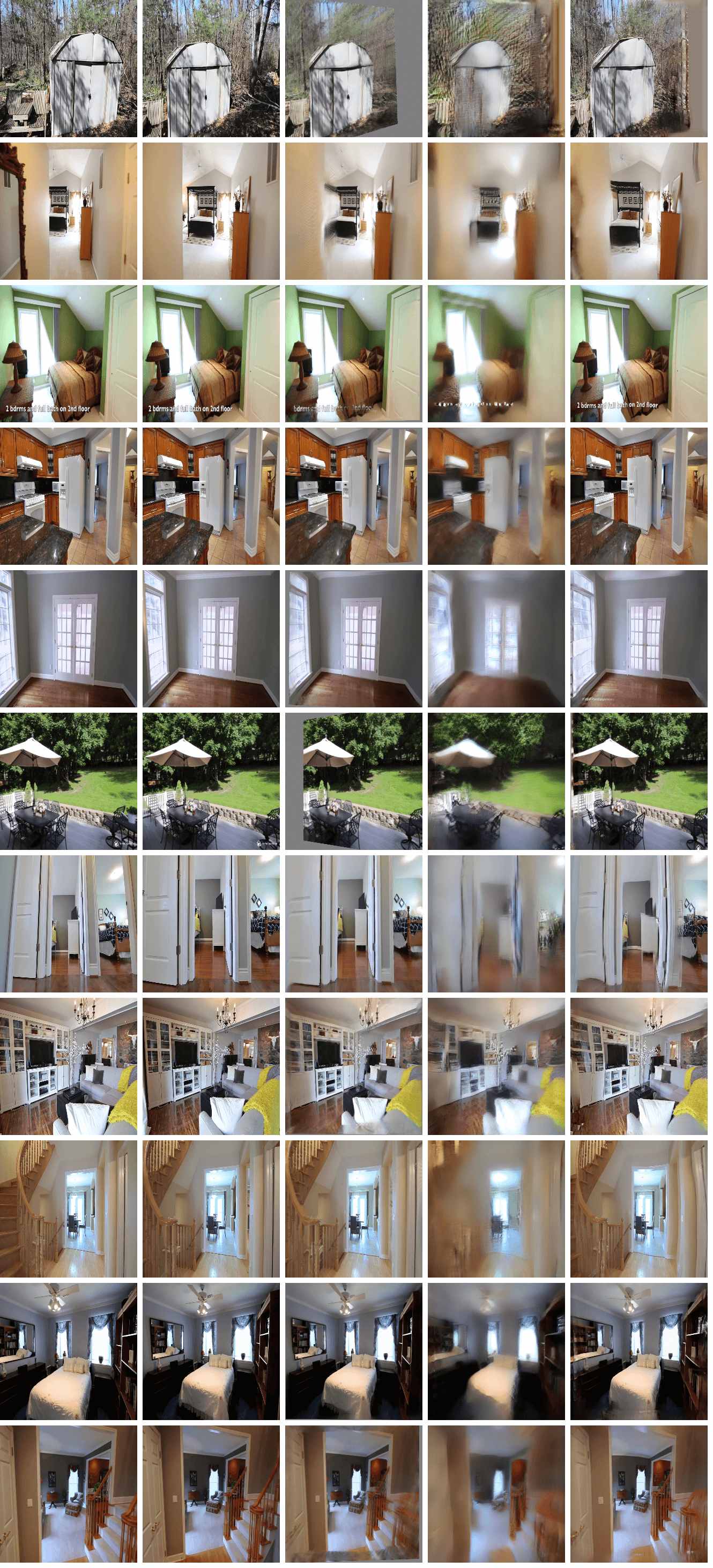}
\scriptsize
\put (1,-1.5) {Input Img}
\put (10,-1.5) {Target Img}
\put (19,-1.5) {StereoMag \cite{Zhou18}}
\put (28,-1.5) {Vox w/ ours}
\put (39,-1.5) {\modelname{}}
\end{overpic}
\vspace{2em}
\caption{Additional results on RealEstate10K \cite{Zhou18}. Zoom in for details.}
\label{fig:real10k1}
\end{figure}

\begin{figure}
\begin{overpic}[width=\linewidth]{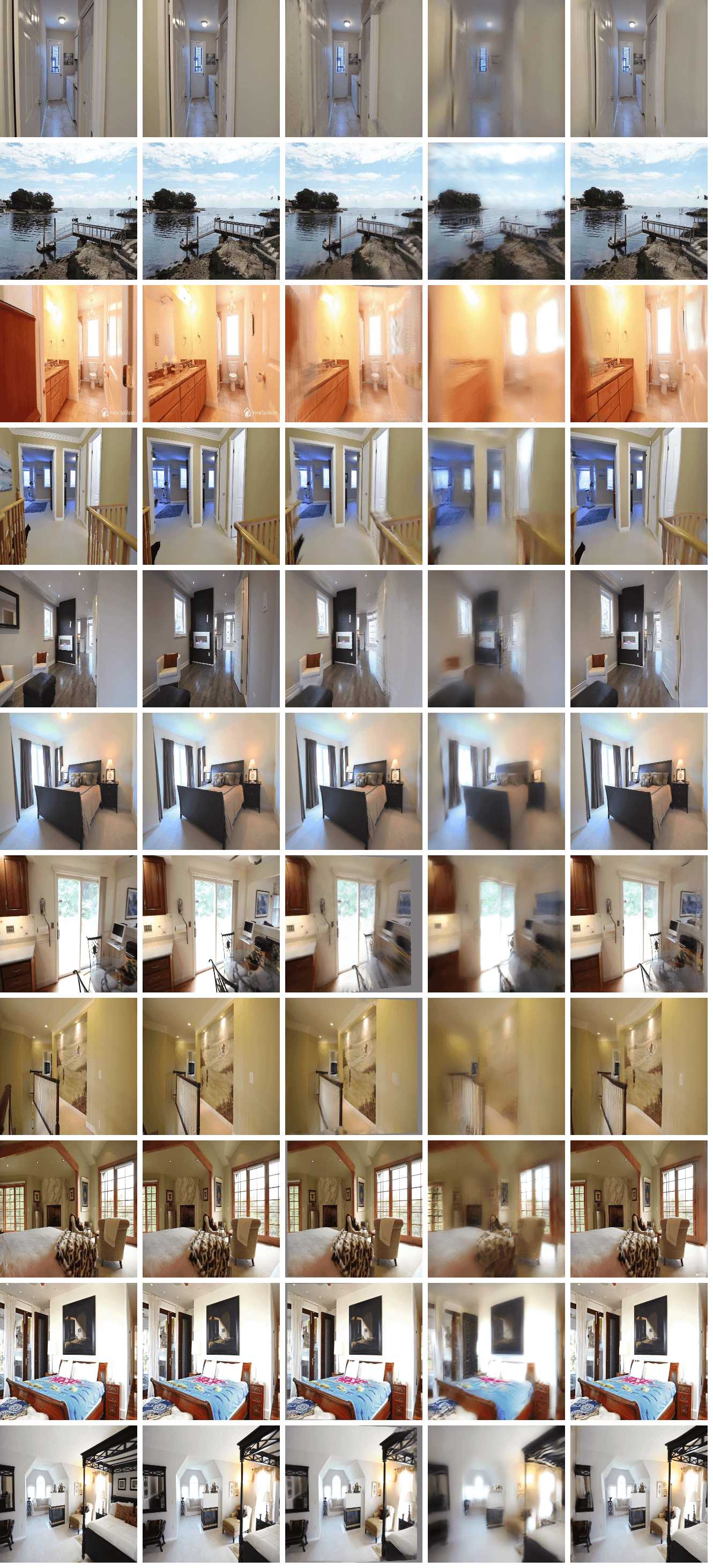}
\scriptsize
\put (1,-1.5) {Input Img}
\put (10,-1.5) {Target Img}
\put (19,-1.5) {StereoMag \cite{Zhou18}}
\put (28,-1.5) {Vox w/ ours}
\put (39,-1.5) {\modelname{}}
\end{overpic}
\vspace{2em}
\caption{Additional results on RealEstate10K \cite{Zhou18}. Zoom in for details.}
\label{fig:real10k2}
\end{figure}

\begin{figure}
\begin{overpic}[width=\linewidth]{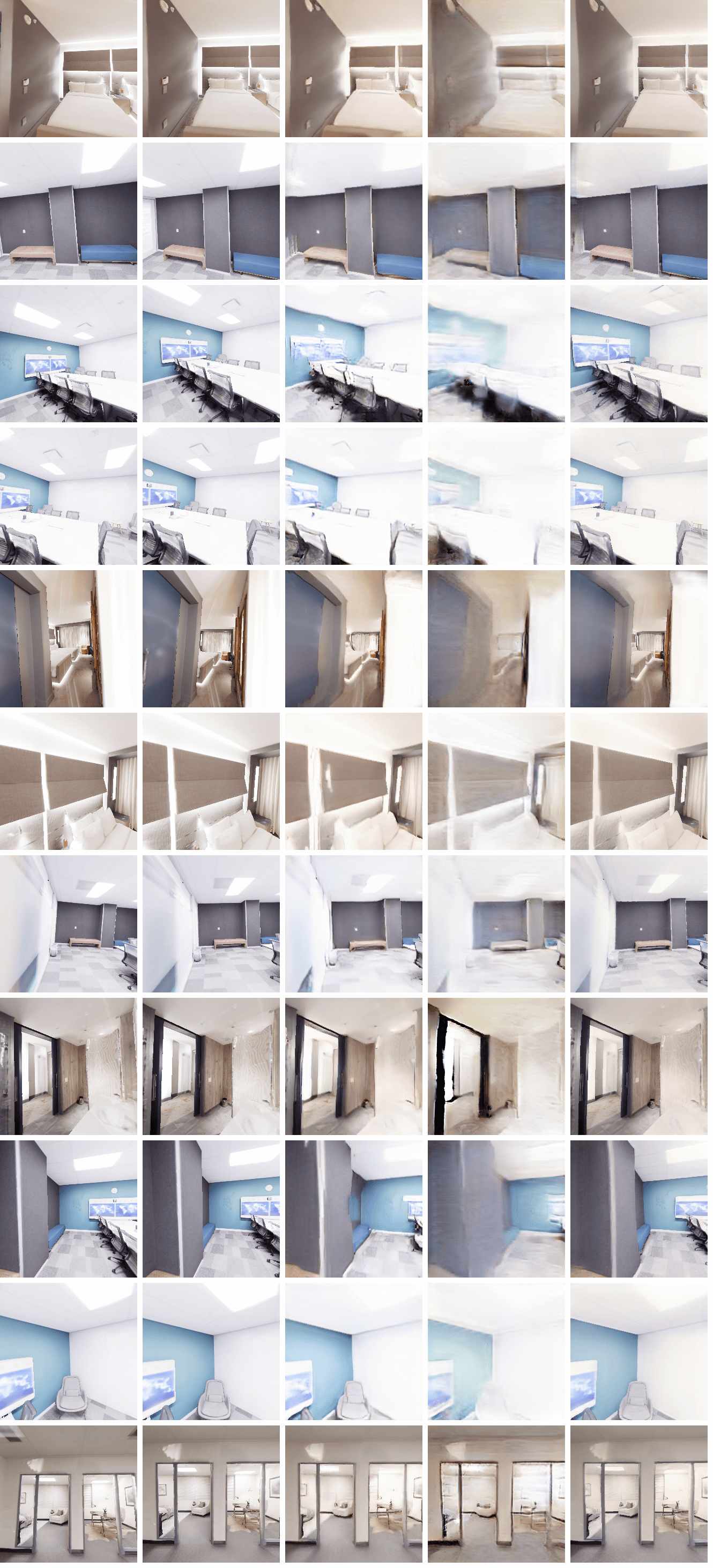}
\scriptsize
\put (1,-1.5) {Input Img}
\put (10,-1.5) {Target Img}
\put (19,-1.5) {Vox w/ ours}
\put (28,-1.5) {Vox w/ unet}
\put (39,-1.5) {\modelname{}}
\end{overpic}
\vspace{2em}
\caption{Additional results on Replica \cite{Replica19}. Zoom in for details.}
\label{fig:replica}
\end{figure}

\begin{figure}
\begin{overpic}[width=\linewidth]{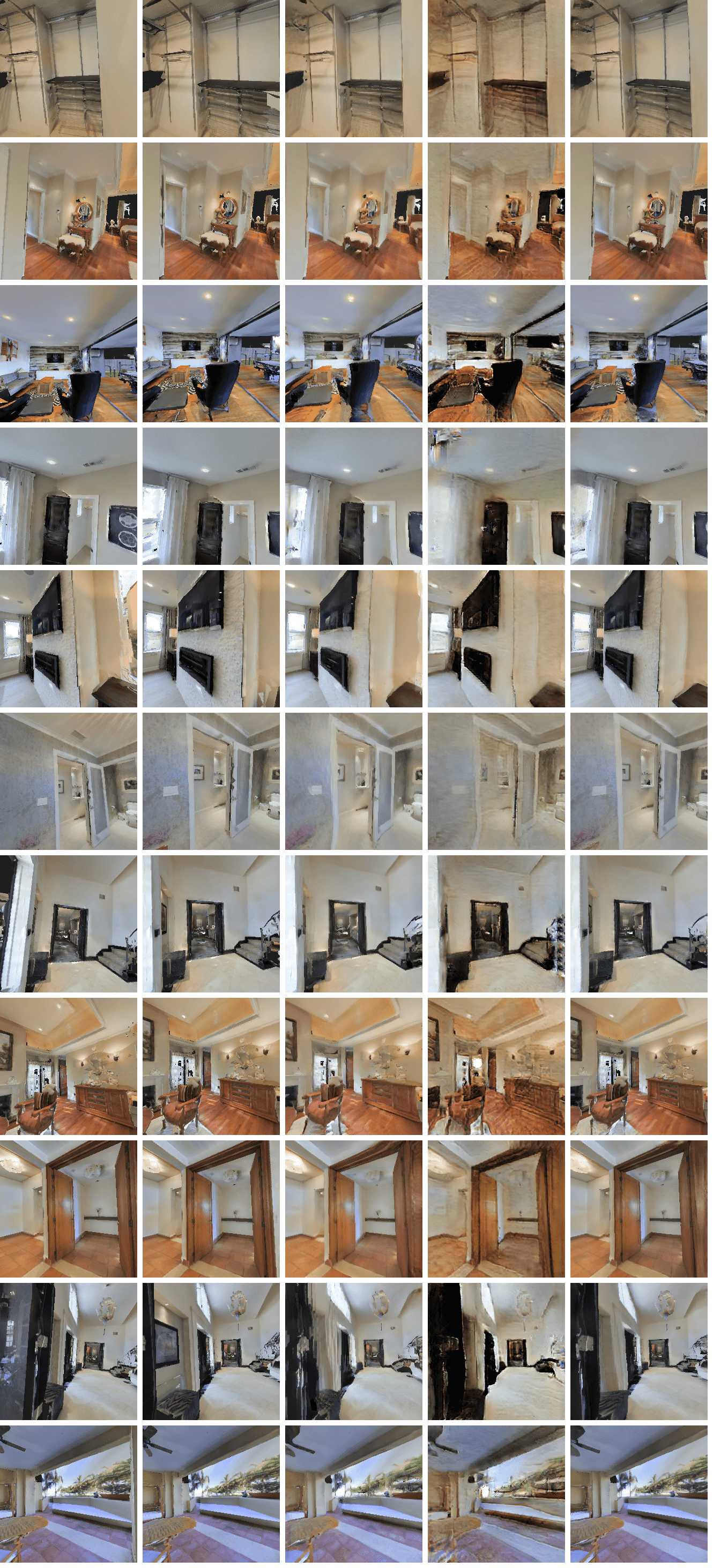}
\scriptsize
\put (1,-1.5) {Input Img}
\put (10,-1.5) {Target Img}
\put (19,-1.5) {Vox w/ ours}
\put (28,-1.5) {Vox w/ unet}
\put (39,-1.5) {\modelname{}}
\end{overpic}
\vspace{2em}
\caption{Additional results on Matterport3D \cite{Matterport3D}. Zoom in for details.}
\label{fig:mp3d}
\end{figure}

\begin{figure}
\centering
\begin{overpic}[width=0.9\linewidth]{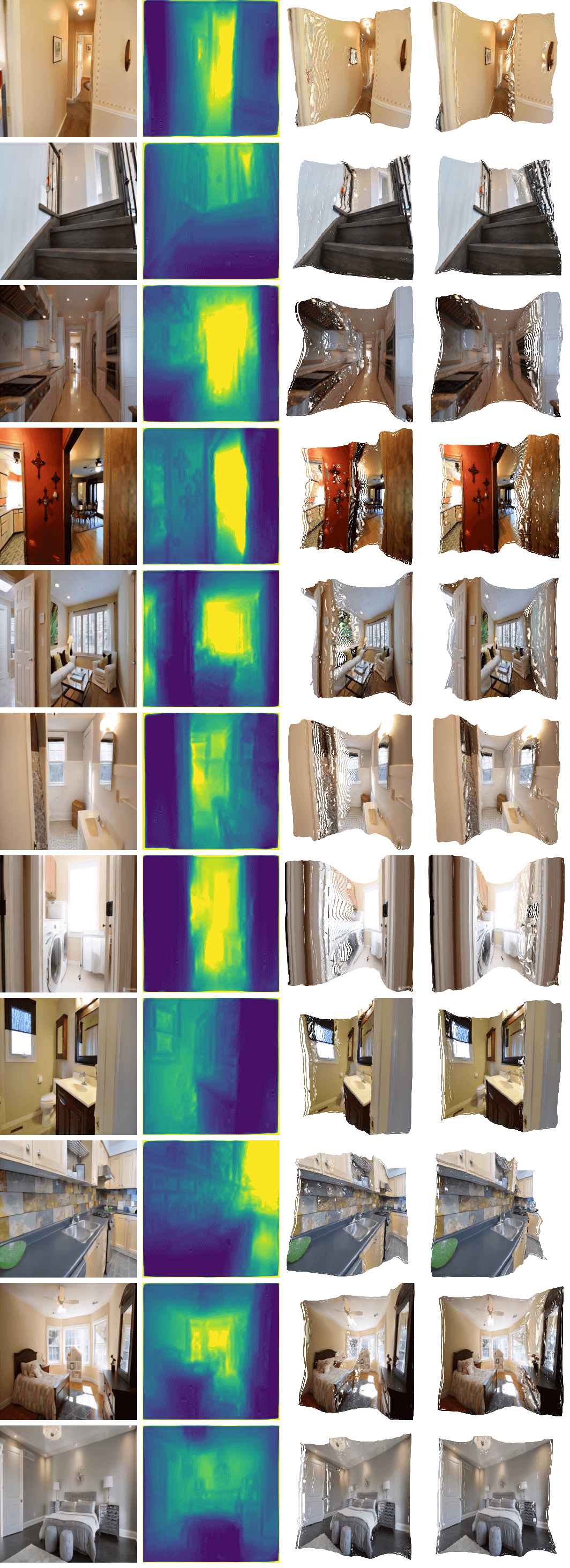}
\scriptsize
\put (2,-1.5) {Input Img}
\put (12,-1.5) {\modelname{}}
\put (18,-1.5) {\modelname{} (PC/$-45^{\circ}$)}
\put (28,-1.5) {\modelname{} (PC/$0^{\circ}$)}
\end{overpic}
\vspace{2em}
\caption{Additional depth predictions on RealEstate10K \cite{Zhou18}. 
We also visualise the point cloud (PC) and the rotated point cloud at $-45^{\circ}$. (Note that the point cloud
in the model is actually a point cloud of features, not RGB values.) }
\label{fig:dep1}
\end{figure}

\begin{figure}
\centering
\begin{overpic}[width=0.9\linewidth]{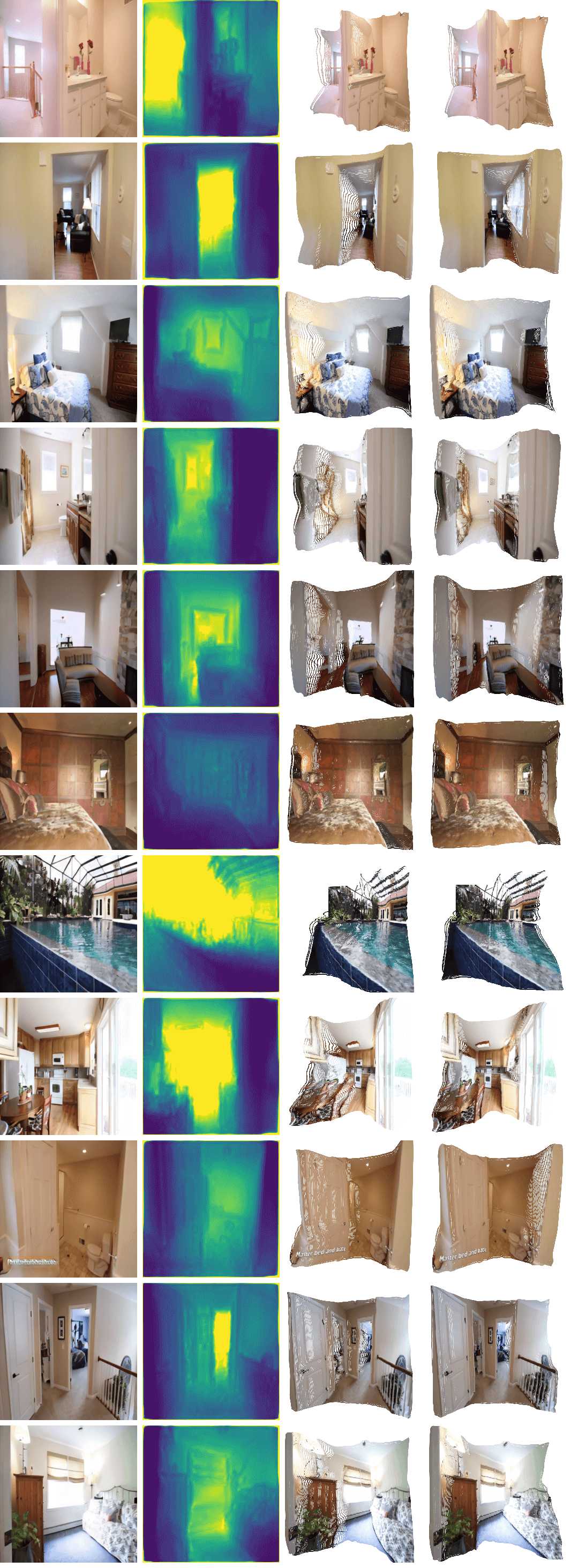}
\scriptsize
\put (2,-1.5) {Input Img}
\put (12,-1.5) {\modelname{}}
\put (18,-1.5) {\modelname{} (PC/$-45^{\circ}$)}
\put (28,-1.5) {\modelname{} (PC/$0^{\circ}$)}
\end{overpic}
\vspace{2em}
\caption{Additional depth predictions on RealEstate10K \cite{Zhou18}.
We also visualise the point cloud (PC) and the rotated point cloud at $0^{\circ}$. (Note that the point cloud
in the model is actually a point cloud of features, not RGB values.)  }
\label{fig:dep2}
\end{figure}

\noindent {\bf Additional qualitative results.} 
We give additional qualitative results on RealEstate10K (\figref{fig:real10k1}-\ref{fig:real10k2}), Replica (\figref{fig:replica}), and Matterport3D (\figref{fig:mp3d}).
The supplementary video shows sample videos of a model generating images along a given trajectory.
We compare \modelname{} to the baseline (Vox w/ ours); \modelname{} has smoother motion with fewer artefacts.
We also visualise additional depth prediction results in \figref{fig:dep1}-\ref{fig:dep2}.

\begin{table}[h]
\centering
\begin{tabular}{lccc}
\toprule
    \multicolumn{4}{c}{\bf Comparison on KITTI \cite{Geiger13}} \\
\cmidrule{1-4}
   &  PSNR $\uparrow$ & SSIM $\uparrow$ & Perc Sim $\downarrow$ \\ 
\cmidrule{2-4}
\modelname{} & {16.70} &  {0.52} &  {\bf 2.07}  \\
\cmidrule{2-4}
ContView \cite{Chen19} 	& {\bf 16.90} & \bf{0.54} & 2.21  \\
\bottomrule
\end{tabular}
\caption{Comparison on KITTI to \cite{Chen19}. $\uparrow$ denotes higher is better, $\downarrow$ lower is better.}
\label{tab:kitti}

\end{table}

\clearpage

\clearpage

\section{Additional architectural details}
\label{sec:archresults}

Here we give more information about the precise architectural details used to build the
components of our model.

\begin{figure*}
\centering
\subfigure[][ResNet block.]{\includegraphics[width=0.35\linewidth]{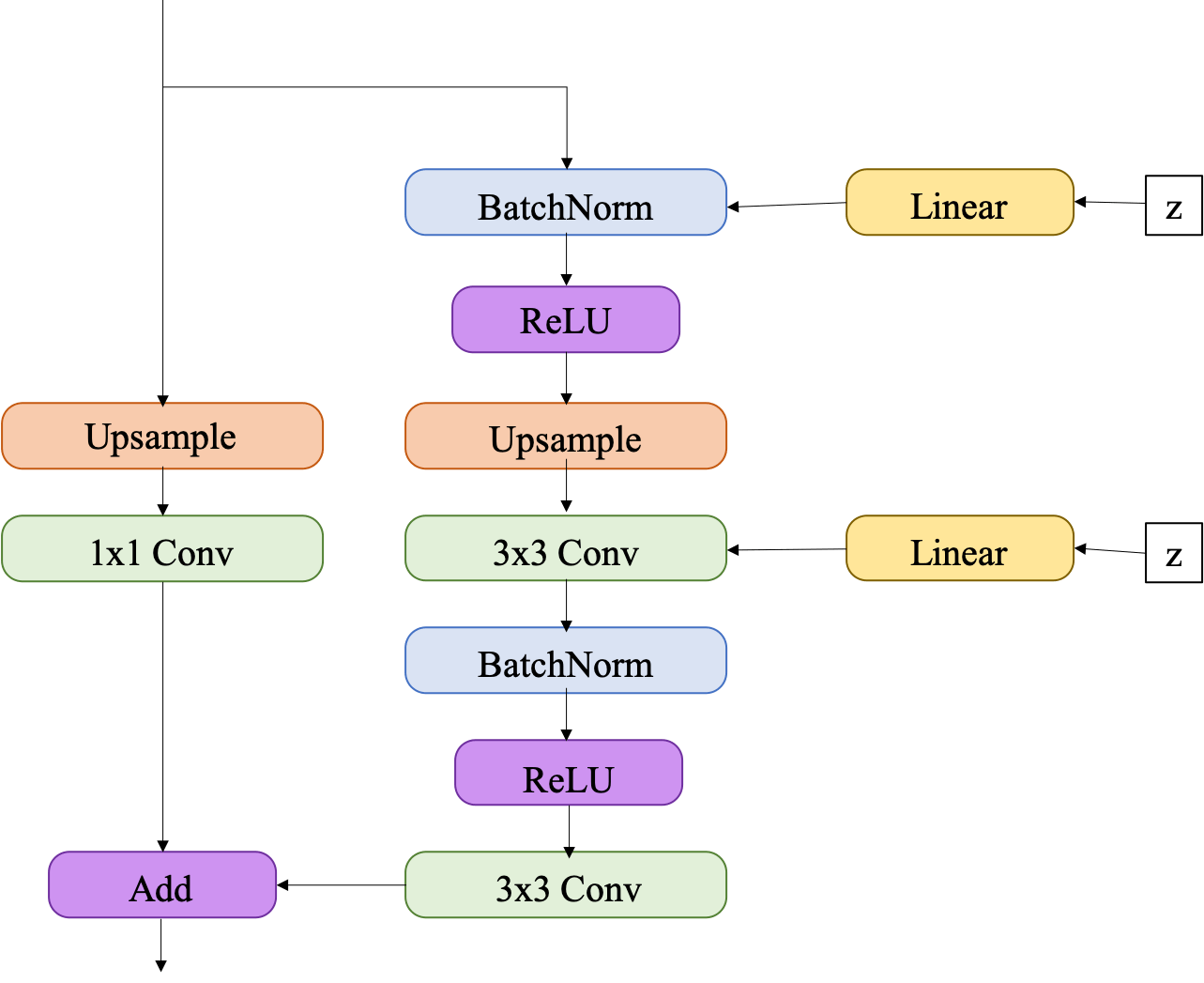}\label{fig:resnetups}} \quad \quad
\subfigure[][ResNet block with an average pool block.]{\includegraphics[width=0.35\linewidth]{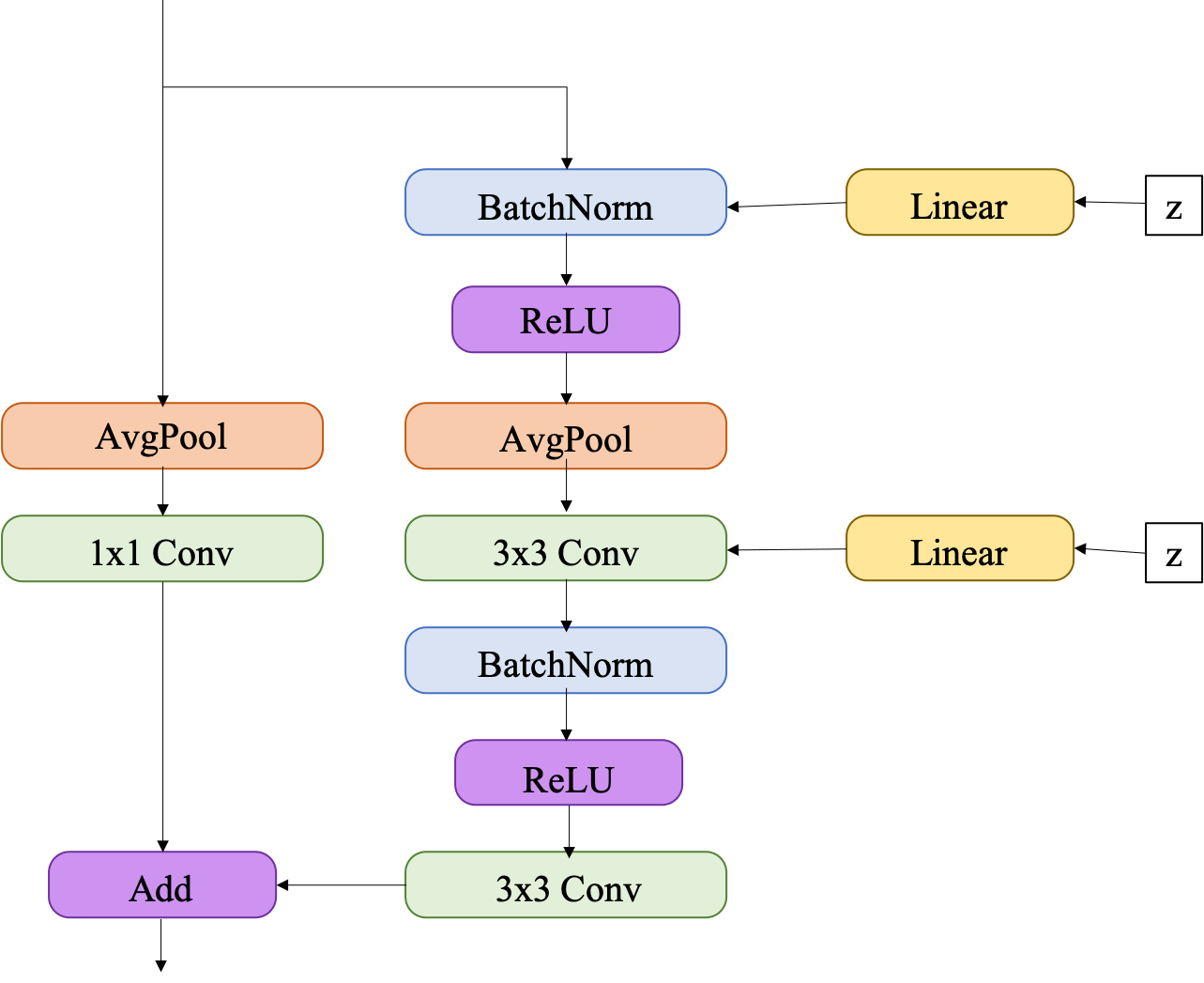}\label{fig:resnetavg}} \quad \quad
\subfigure[][ResNet block with an identity block.]{\includegraphics[width=0.35\linewidth]{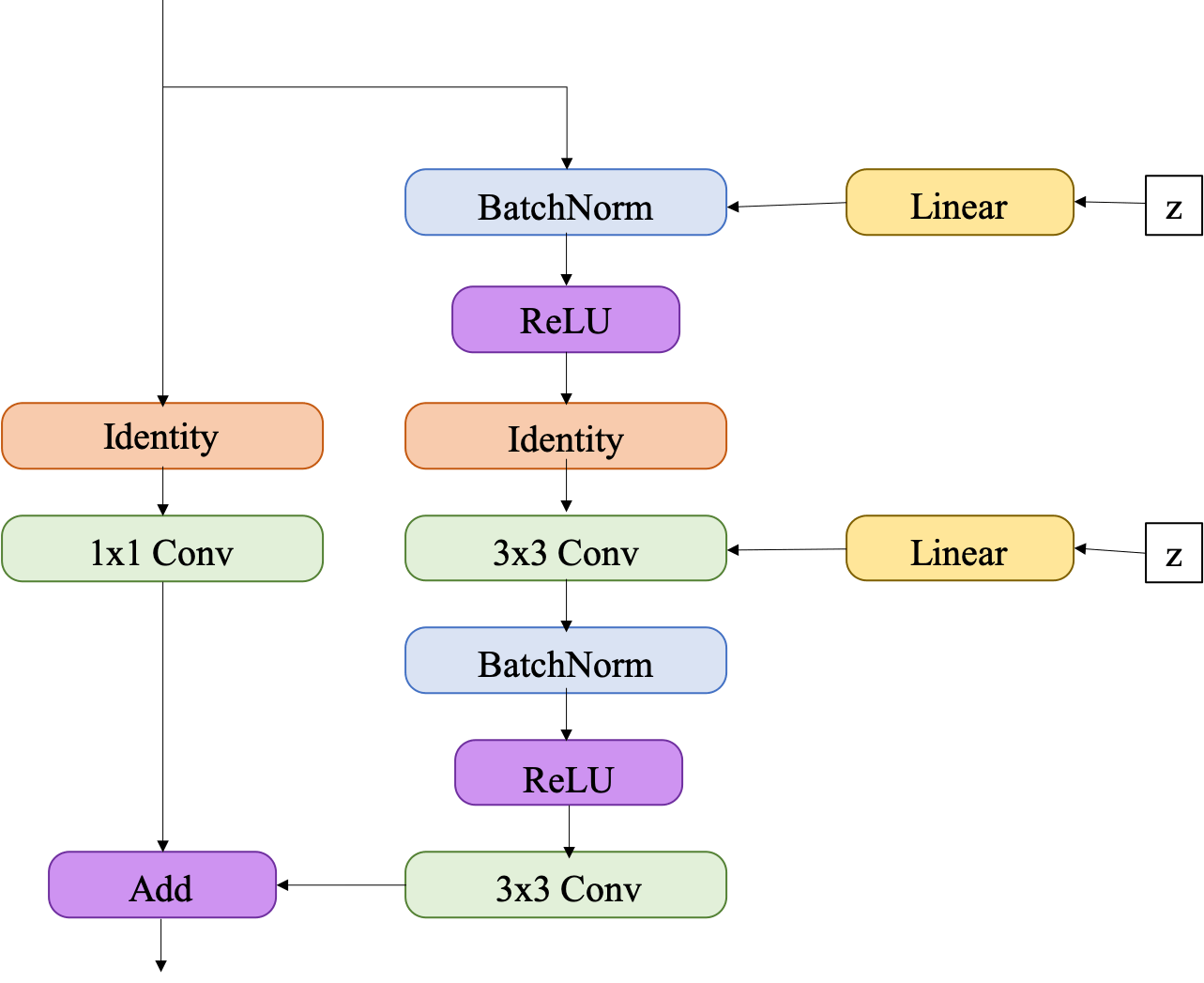}\label{fig:resnetid}}
\caption{An overview of ResNet blocks. In (a), we show the basic ResNet block, 
(b) when we replace the upsample block by an average pool block, and 
(c) when we replace the upsample block by an identity block.}
\label{fig:resnetblocks}
\end{figure*}

\paragraph{\bf ResNet blocks.}
Our spatial feature network and refinement networks are composed of ResNet blocks.
The ResNet blocks used are the same as those used in \cite{Brock19} (Appendix B, Fig 15 (b)), reproduced in \figref{fig:resnetblocks}.
However, we consider three different setups.
The block may be used to increase the resolution of the features using an upsample layer (as used in the original paper by \cite{Brock19}) (\figref{fig:resnetups}).
The block may be used to decrease the resolution of the features using an average pooling layer as opposed to the upsample layer (\figref{fig:resnetavg}).
The block may be used to maintain the resolution of the features using an identity layer as opposed to the upsample layer (\figref{fig:resnetid}).

\begin{figure}
\centering
\subfigure[][Spatial feature network.]{\includegraphics[height=0.8\linewidth]{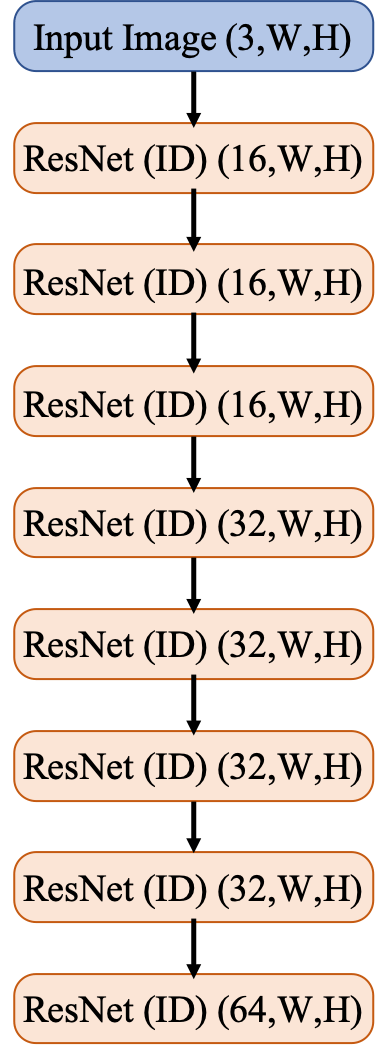}\label{fig:encoder}} \quad \quad
\subfigure[][Refinement network.]{\includegraphics[height=0.8\linewidth]{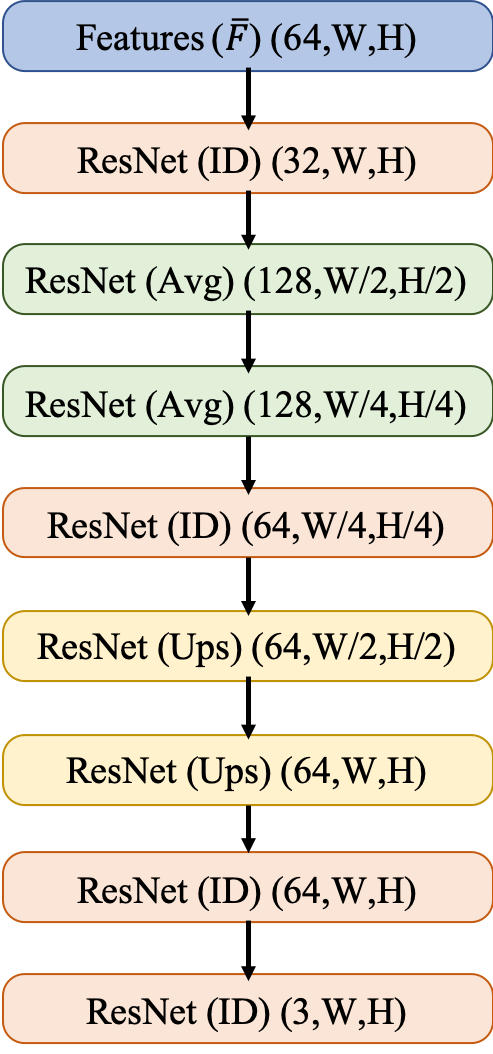}\label{fig:decoder}}
\caption{Our sequence of ResNet blocks in the spatial feature and refinement networks.}
\end{figure}

\paragraph{\bf Spatial feature network.}
ResNet blocks are stacked together to form the embedding network.
In particular, we use the setup in \figref{fig:encoder}.

\paragraph{\bf Refinement network.}
ResNet blocks are stacked together to form the decoder network.
In particular, we use the setup in \figref{fig:decoder}.

\begin{figure}
\centering
\includegraphics[height=1.4\linewidth]{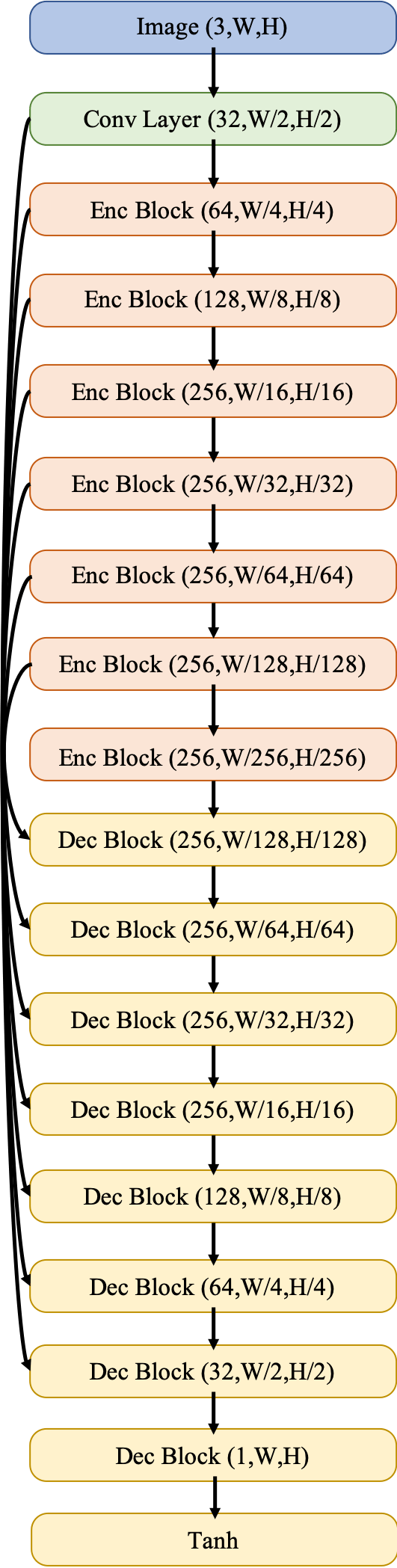}
\caption{Depth regressor network. An {\em Enc Block} consists of a sequence of Leaky ReLU, convolution (stride 2, padding 1, kernel size 4), and batch normalisation layers.
A {\em Dec Block} consists of a sequence of ReLU, 2x bilinear upsampling, convolution (stride 1, padding 1, kernel size 3), and batch normalisation layers (except for the final layer, which has no batch normalisation layer). }
\label{fig:unet}
\end{figure}

\paragraph{\bf Depth regressor.}
The depth regressor network uses a UNet architecture, as illustrated in \figref{fig:unet}.

\paragraph{\bf Additional details on the perceptual loss.}
We follow the perceptual loss used in \cite{Park19}.

\section{Additional details on baselines}
\label{sec:baselines}

In this section, we give further information about the baselines used.

\begin{figure}
\centering
\includegraphics[height=1.4\linewidth]{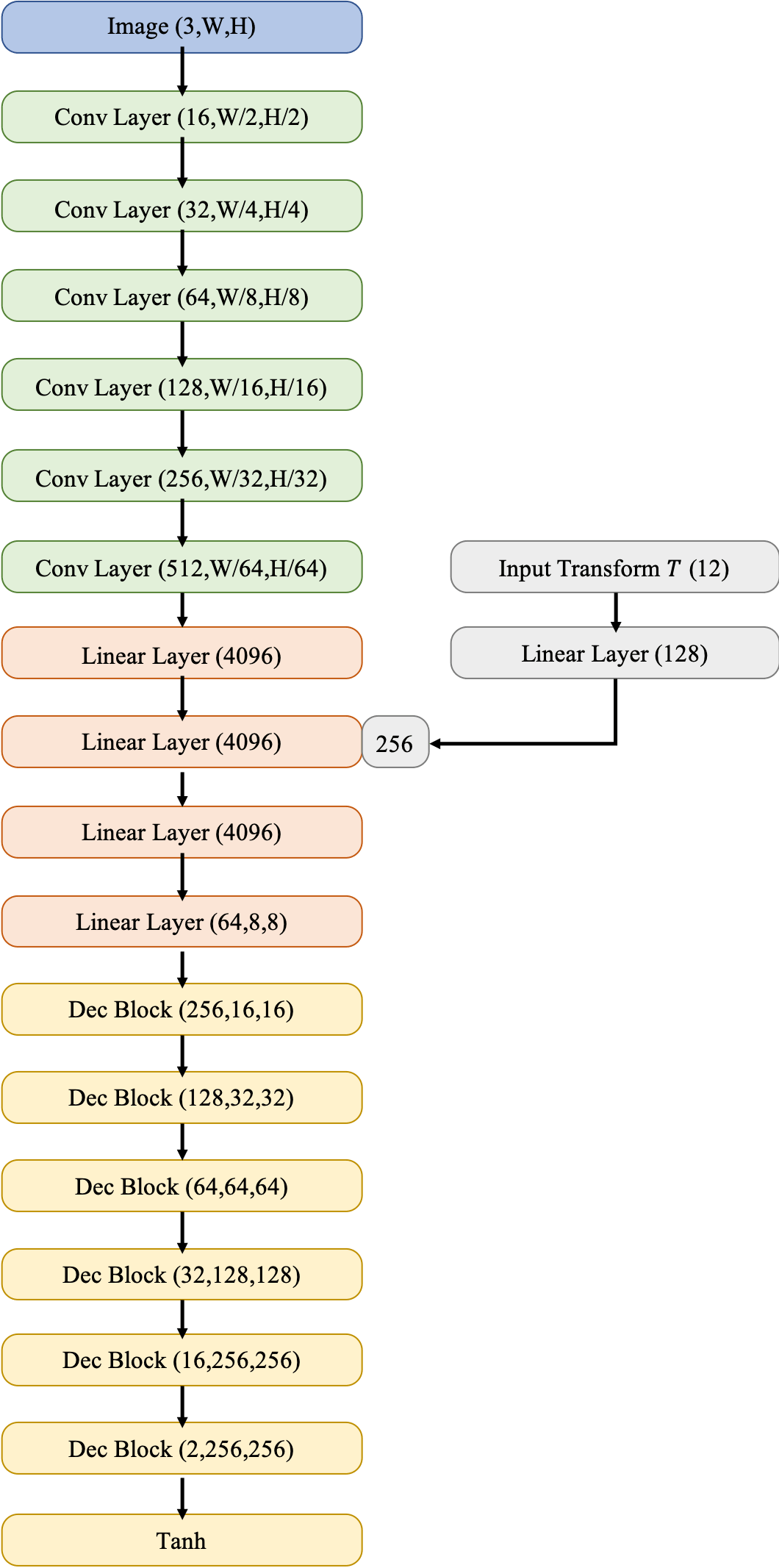}
\caption{An overview of the image to image network. A {\em Conv Layer} consists of a sequence of a convolutional layer (stride 2, padding 1, filter size 3), ReLU, and batch normalisation layer. 
A {\em Linear Layer} consists of a sequence of a linear layer, ReLU, and batch normalisation layer. A {\em Dec block} consists of a sequence of a convolutional layer (stride 1, padding 1, filter size 3), ReLU, batch normalisation layer and upsample layer (except for the last, which consists of simply a convolutional layer).}
\label{fig:im2im}
\end{figure}

\paragraph{\bf Im to im.}
\label{sec:im2im}
We follow the architecture of \cite{Zhou16}.
However, \cite{Zhou16} only considers discrete rotations about the azimuth and a small set of changes in elevation, so
\cite{Zhou16} takes four values as input, the $\cos$ and $\sin$ values of the azimuth and elevation.
 However, our datasets include rotation in all three directions, as well as translational motion.
 As a result, we modify their angle encoder to take 12 values (as opposed to four), and pass the change in viewpoint, $\m T$
 to the angle encoder. 
 The network is visualised in \figref{fig:im2im}.

\paragraph{\bf Vox w/ unet.}
\label{sec:voxunet}
This baseline is based on \cite{Sitzmann19}, which represents 3D shape in a neural network using a voxel representation.
Note that they train one model per instance, so their model only generalises to that one object. 
Their overall setup is as follows.
An image is passed through an encoder (\eg our spatial feature network) to obtain a set of features.
The features are projected into a voxel grid, which is transformed and projected into the new view.
The features are accumulated using an occlusion network, which acts as a pseudo depth predictor and predicts the occupancy of the voxels.
The predicted occupancy is used to re-weight and combine features.
This is then passed to the decoder (\eg our refinement network) which predicts the scene at the new view.
Finally, the generated image is compared to the true image using discriminators and photometric losses.

\begin{figure}
\centering
\subfigure[][Encoder network.]{\includegraphics[height=1.3\linewidth]{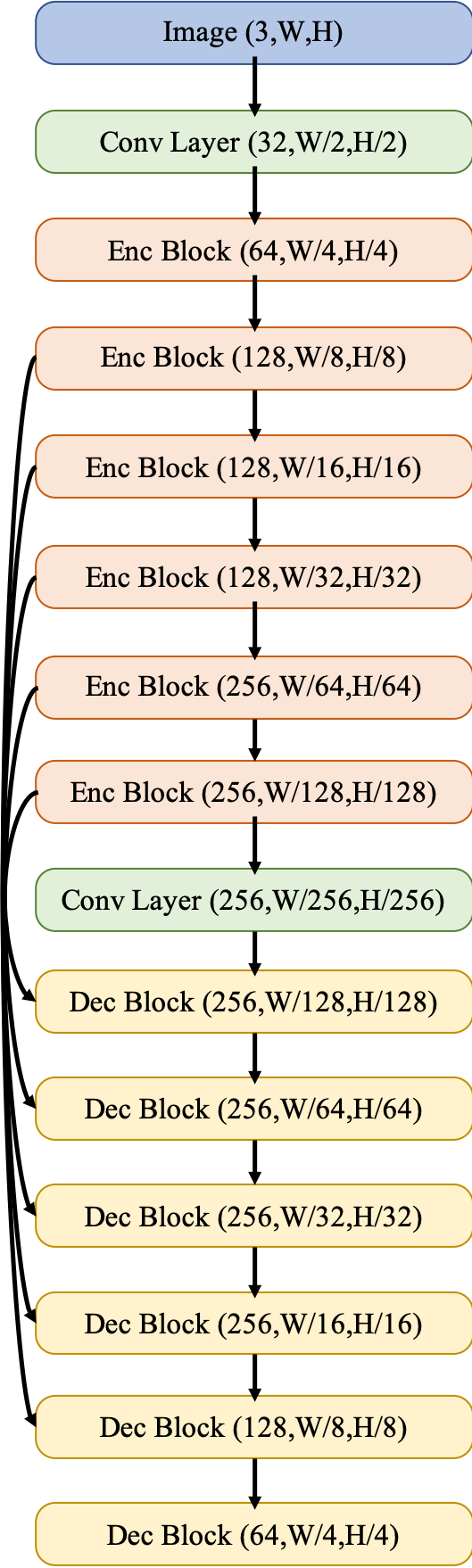}} \quad \quad
\subfigure[][Decoder network.]{\includegraphics[height=1.3\linewidth]{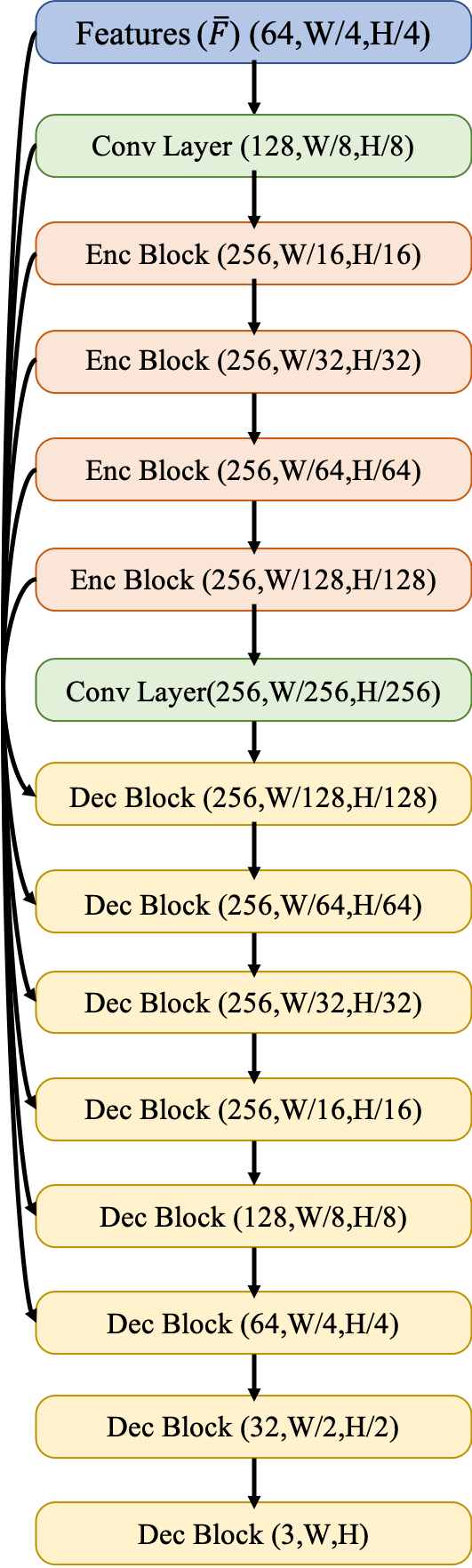}}
\caption{The encoder and decoder network for the UNet style encoder/decoder setup. An {\em Enc block} is a sequence of a LeakyReLU, convolutional layer (stride 2, padding 1, kernel size 4) and
batch normalisation layer. A {\em Dec block} is a sequence of ReLU, bilinear upsampling layer, convolutional layer (stride 1, padding 1, kernel size 3), and batch normalisation layer (except for the last layer which has no batch normalisation).}
\label{fig:voxunet}
\end{figure}

\begin{figure}
\centering
\includegraphics[height=1.3\linewidth]{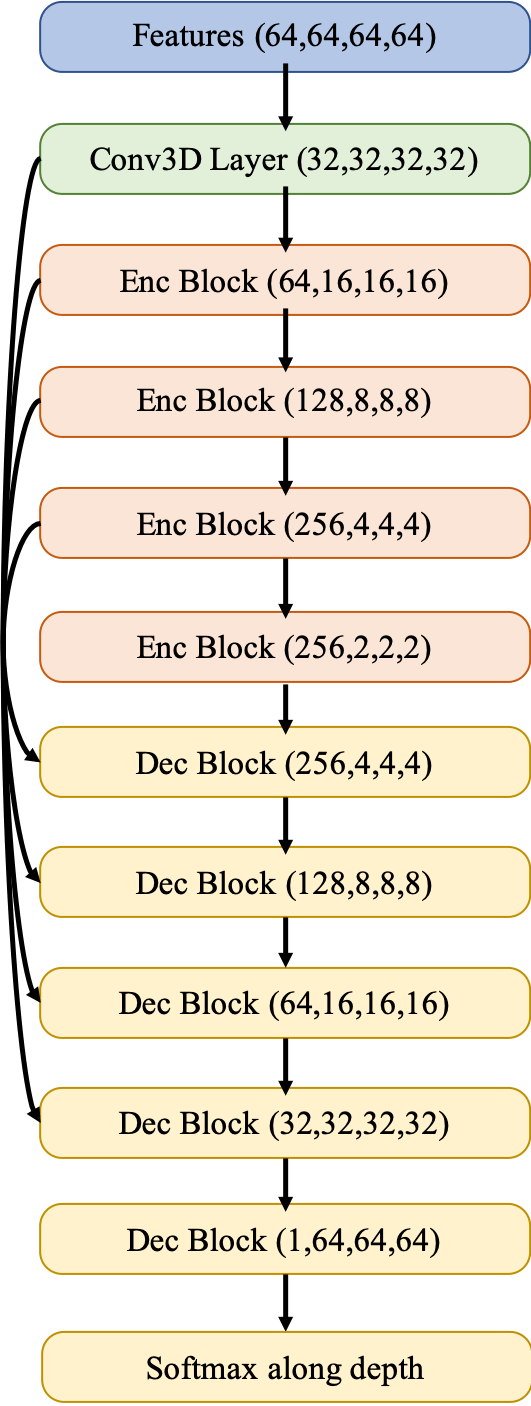}
\caption{The 3D UNet for predicting the occupancy of voxels. An {\em Enc block} consists of a sequence of a LeakyReLU, convolutional layer (stride 2, padding 1, kernel size 4) and
batch normalisation layer. A {\em Dec block} consists of a sequence of ReLU, bilinear upsampling layer, convolutional layer (stride 1, padding 1, kernel size 3), and batch normalisation layer (except for the last layer which has no batch normalisation).}
\label{fig:voxunet3d}
\end{figure}

To reimplement this approach, we follow their architectural choices and use a UNet style architecture for all network components (the spatial feature network, refinement network,
and occlusion network). However, we use the discriminators and photometric losses used to train \modelname{} to ensure that both methods are
fair in terms of the discriminator.
The details for the encoder/decoder setup are given in \figref{fig:voxunet}.
The occupancy network is a 3D UNet, which takes as input the rotated voxels and then predicts occupancy for each voxel location; these are
then normalised using a softmax layer over the depth dimension.
The details are given in \figref{fig:voxunet3d}.
We use their setup but train the network to generate new images of a scene given a {\em single} image of a scene.

\begin{figure}
\centering
\subfigure[][Encoder network.]{\includegraphics[height=0.8\linewidth]{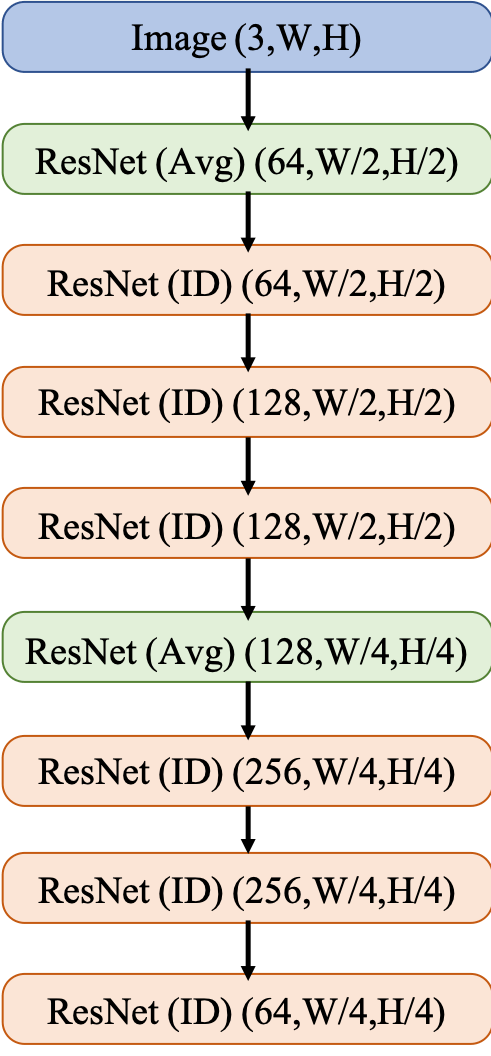}\label{fig:unetencoder}} \quad \quad
\subfigure[][Decoder network.]{\includegraphics[height=0.8\linewidth]{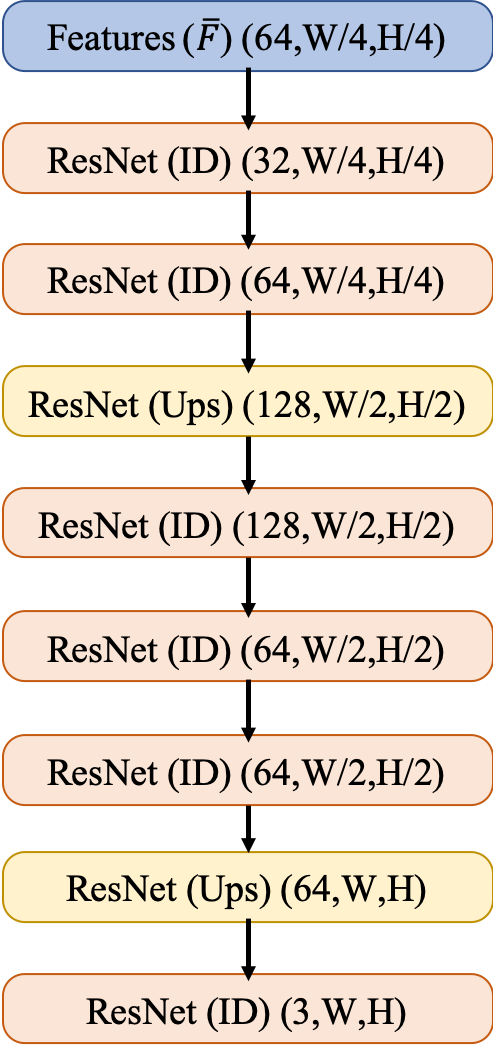}\label{fig:unetdecoder}}
\caption{The spatial feature and refinement networks for the ResNet style setup in the Vox w/ ours baseline.}
\label{fig:unetours}
\end{figure}

\paragraph{\bf Vox w/ ours.}
Instead of using the UNet style spatial feature and refinement network in vox w/ ours, we use a sequence of ResNet blocks,
as described in \figref{fig:unetours}.
The set of ResNet blocks in the spatial feature network downsamples the image to the appropriate size.
The refinement network similarly upsamples the projected features to the appropriate image size.
We also use a larger capacity in this setup to ensure that our 3D representation is preferable.
The network was trained with a lower learning rate (lr=0.0004)  as opposed to (lr=0.001) as in our model, as 
we found that the model struggled to learn with the higher learning rate.

{\bf Other setups.} We experimented with other ResNet block sequences and multiple learning rates when creating this baseline.
Instead of downsampling the features within the encoder (\eg the spatial feature network), we can use the same spatial feature network as \modelname{} (to obtain features of size $C\times256\times256$ and then downsample to obtain features of size $C\times64\times64$.
Similarly, instead of upsampling the features within the decoder (\eg the refinement network), we can upsample the transformed features to obtain ones of size $C\times256\times256$ and pass these upsampled
features to the refinement network and so use the same refinement network we use in \modelname{}.
We found that the results were similar to those of the model used in the paper on RealEstate10K but worse on Matterport.

We additionally found that the results were highly dependent on the learning rate for this model.

\paragraph{\bf 3DView.}
This baseline is based on a depth predictor (\eg \cite{Li18}), so 3DView predicts depth up to a scale ambiguity.
As the depth is only predicted up to a scale, we generate images for multiple possible scales for each test image and then report results for the
best image.

\clearpage

\section{Additional information about datasets}
\label{sec:datasets}

\paragraph{\bf Matterport3D. }
For Matterport, the minimum depth  is $0.1$ and the maximum depth $10$.

\paragraph{\bf RealEstate10K. }
For RealEstate10K, the minimum depth is $1$ and the maximum depth is  $100$.

\paragraph{\bf KITTI. }
For KITTI, the minimum depth is $1$ and the maximum depth is  $50$.

\section{A description of other setups we tried}
\label{sec:negativeresults}
\paragraph{\bf Model setup}
\begin{itemize}
\item We experimented with using a UNet architecture instead of a sequence of ResNet blocks for the spatial feature network and refinement network.
This led to much worse results and was more challenging to train.
\end{itemize}

\paragraph{\bf Differentiable renderer setup}
\begin{itemize}
\item Other settings for the differentiable renderer: We tried a larger radius, $r=8$, but this both takes longer to train and gives worse results.

\item Other settings for the accumulation function: We tried using a weighted sum with and without normalisation for the accumulation step.
These led to similar results, but without normalisation had noisier training characteristics. The implementation of these different accumulation
setups is available in the online code. 
\end{itemize}

\end{document}